%% file: probOpt.tex
\documentclass[letterpaper,10pt]{article}
\usepackage[T1]{fontenc}
\usepackage{parskip}
\usepackage[font={small}]{caption}
\usepackage[hidelinks]{hyperref}
\usepackage[margin=2.5cm]{geometry}
\usepackage{times}
\usepackage{amsmath,amssymb,amsthm}
\usepackage[affil-it]{authblk}

\usepackage{graphicx} 
\usepackage{times} 
\usepackage{amsmath} 
\usepackage{amssymb}  
\usepackage{color}
\usepackage{caption}
\usepackage{subcaption}

\usepackage{esvect}

\graphicspath{{./Figures/}}

\usepackage{algorithm,algorithmic,url,listings}

\newcommand{\Exp}[1]{\operatorname{E}\left[#1\right]}

\newcommand{\vecV}[1]{\operatorname{vec}\left(#1\right)}
\newcommand{\vecVh}[1]{\operatorname{vech}\left(#1\right)}

\renewcommand\mid{\,\vert\,}

\newcommand{\T}{N}
\newcommand{\R}{\mathbb{R}}

\newcommand{\N}{\mathcal{N}}

\newcommand{\GP}{\mathcal{GP}}

\newcommand{\Transp}{\mathsf{T}}

\newcommand{\bmat}[1]{\begin{bmatrix}#1\end{bmatrix}}

\newcommand{\nTh}{n_{\theta}}              
\newcommand{\nX}{n_{x}}                        
\newcommand{\nY}{n_{y}}                        
\newcommand{\setX}{\ensuremath{\mathsf{X}}}                      
\newcommand{\setY}{\ensuremath{\mathsf{Y}}}                      
\newcommand{\setTh}{\ensuremath{\mathsf{\Theta}}}           

\newcommand{\myd}{\textrm{d}}      

\date{}

\title{On the construction of probabilistic Newton-type algorithms}
\author[1]{Adrian G. Wills}
\author[2]{Thomas B. Sch\"on}
\affil[1]{School of Engineering, University of Newcastle,
  Australia. 
E-mail: adrian.wills@newcastle.edu.au}
\affil[2]{Department of Information
    Technology, Uppsala University, Sweden. 
E-mail: thomas.schon@it.uu.se}

\begin{document}

\maketitle

%
%
%

\begin{abstract}
  %
  It has recently been shown that many of the existing quasi-Newton
  algorithms can be formulated as learning algorithms, capable of
  learning local models of the cost functions. Importantly, this
  understanding allows us to safely start assembling probabilistic
  Newton-type algorithms, applicable in situations where we only have
  access to noisy observations of the cost function and its
  derivatives. This is where our interest lies. 
  %
  We make contributions to the use of the non-parametric and
  probabilistic Gaussian process models in solving these stochastic
  optimisation problems. Specifically, we present a new algorithm that unites these
  approximations together with recent probabilistic line search routines to
  deliver a probabilistic quasi-Newton approach.  
  %
  We also show that the probabilistic optimisation algorithms deliver
  promising results on challenging nonlinear system identification
  problems where the very nature of the problem is such that we can
  only access the cost function and its derivative via noisy
  observations, since there are no closed-form expressions available.
\end{abstract}


\section{Introduction}
We are in this paper concerned with the unconstrained nonlinear
optimisation problem
\begin{align}
  \label{eq:21}
  \widehat{x} = \arg \min_{x}{f(x)},
\end{align}
in situations when we only have access to \emph{noisy} evaluations of
the cost function~$f(x)$ and its derivatives. In the noise-free case,
solving this problem has attracted enormous research attention for
many decades and this has resulted in many variants of optimisation
algorithms. Among these are the much celebrated class of quasi-Newton
methods that are still---almost half a century after their
inception---the state of the art methods when it comes to numerical
solution of the unconstrained optimisation problem~\eqref{eq:21}. This
is true across most---if not all---branches of science where
optimisation problems of the type~\eqref{eq:21} needs to be solved.

When $f(x)$ and its derivatives are noisy, then these existing methods
suffer from a fundamental problem. Specifically, the algorithms may
fail to converge since they rely on knowledge of descent directions
and line-search (or related) methods that are not geared towards noisy
cost functions and gradient evaluations. Such stochastic optimisation
problems are commonly occurring for example in the following
situations; 1) If the dataset is very large it is not possible to
evaluate the cost function on the entire dataset and instead it is
divided into so-called \emph{mini-batches}. This situation arises
quite often in Machine Learning and in particular in deep learning
applications.  2) When we employ numerical algorithms to compute the
cost function and its derivatives. This occurs for example in
nonlinear system identification using the maximum likelihood method
when various particle filters are used to compute the intractable cost
function and its gradients. We use nonlinear system identification as
a case study in this paper.

There has recently been some very relevant and encouraging
developments for dealing with these stochastic optimisation
problems. More specifically it has been
shown~\cite{HennigK:2013,Hennig:2015} that standard quasi-Newton
methods like the BFGS
method~\cite{Broyden:1970,Fletcher:1970,Goldfarb:1970,Shanno:1970},
Broyden's method~\cite{Broyden:1965} and the DFP
formula~\cite{FletcherP:1963,Broyden:1967} all can be interpreted as
particular instances of Bayesian linear regression or as Gaussian
process regression. This line of research has shown that we can
reinterpret the quasi-Newton algorithms as learning algorithms that
estimate a local quadratic model to the cost
function~\cite{HennigK:2013,Hennig:2013}. Most of this recent
development has taken place within the relatively new and vibrant
direction of research commonly referred to as \emph{probabilistic
  numerics}, see \url{probabilistic-numerics.org}
and~\cite{HennigOG:2015}.

Perhaps most importantly this line of research has opened up for
genuinely new probabilistic optimisation algorithms, which is
necessary in order to solve the problems we are facing when we only
have noisy observations of the cost function and its derivatives. We
will in this work explore these ideas and present a new algorithm that
combines a new approach to modelling the Hessian matrix together with
recent results for probabilistic line search routines. Importantly, we are able to ensure that the
model of the Hessian is symmetric by making use of the
half-vector operator~\cite{MagnusN:1980}. Another technical
contribution is that we propagate the uncertainty in the Hessian
approximation between iterations of the algorithm. As a final
contribution we have the application of these new algorithms to the
challenging nonlinear system identification problem.

%
The key construction is provided by the Bayesian non-parametric
Gaussian process (GP)~\cite{MacKay:1998,RasmussenW:2006}, which is
very briefly introduced in Section~\ref{sec:GP}. In that section we
also outline the two main directions of development that exist when it
comes to modeling optimisation problems using the GP. These two
directions are then described and developed further in
Sections~\ref{sec:NPqN} and~\ref{sec:GlobalGP}. The algorithms are
then profiled on the nontrivial nonlinear system identification
application in Section~\ref{sec:experiments}. Finally, we state our
conclusions and ideas for future direction of this research in
Section~\ref{sec:conclusions}.



\section{Modeling optimisation problems using GPs}
\label{sec:GP}

\subsection{Background on the Gaussian process}
%
The Gaussian process~\cite{MacKay:1998,RasmussenW:2006} is by now an
established model for \emph{nonlinear} functions. The representation
that is used by the GP in modeling a nonlinear function is
\emph{non-parametric} (meaning that it does not rely on any parametric
functional form) and \emph{probabilistic} (meaning that uncertainty is
taken into account throughout the model).

%
The Gaussian process is formally defined as a (potentially infinite)
collection of random variables such that any finite subset of it has a
joint Gaussian distribution. Let us assume that we want to model some
nonlinear function~$f(x)$ as a realisation from a Gaussian process. We
then assign a prior distribution over the function $f(x)$ given be the
GP, which we denote by
\begin{align}
  \label{eq:1}
  f(x) \sim \GP (\mu(x), k(x,x^\prime),
\end{align}
where $\mu(x)$ is some suitable mean function (for example, a
strictly convex function centred on prior knowledge of the parameter
values). The covariance function (also referred to as the kernel)
$k(x,x^\prime)$ represents the correlation between function
values based on the two evaluation points~$x$ and~$x^\prime$. 

This prior can then be updated using observations of the function via
the standard results on partitioned Gaussian distributions, see
e.g. \cite{RasmussenW:2006} for details. 

\subsection{Two existing directions}
The idea of using the GP for optimisation is rather natural,
especially in situations where we only have access to noisy
observations of the cost function and its derivatives. The approaches
available so far can very broadly be divided into two directions.

The first direction starts by assuming that the Hessian is distributed
according to a GP. This Hessian is then updated via (potentially
noisy) observations of the Hessian, the gradient and the cost
function. The observations of the gradients and the cost function take
the form of line integral observations of the GP, which can readily be
incorporated. Hennig recently outlined some promising and highly
interesting developments along these
lines~\cite{Hennig:2013,HennigK:2013}. We follow this direction in
Section~\ref{sec:NPqN}.

The second direction instead tries to build a global model of the cost
function and possibly also of its derivatives. Here the crucial
observation is that the derivative of a GP is another
GP~\cite{RasmussenW:2006}. This straightforwardly opens up for the
possibility of modelling the cost function and possibly also its
derivatives as a joint GP. This global GP model is then updated using
the (possibly noisy) observations of the cost function and its
gradients. Developments along this line started in the global
optimisation literature~\cite{Jones:2001} under names such as \emph{Gaussian
  Process Optimisation (GPO)}~\cite{OsborneGR:2009} and \emph{Bayesian
  optimisation}~\cite{ShahriariSWAdF:2016}. This direction is
following and developed in Section~\ref{sec:GlobalGP}.

Interestingly the algorithms resulting from these two directions are
highly suitable for standard nonlinear deterministic problems. We see
great potential in new optimisation algorithms being created not only
for the stochastic situation, but also for the classic deterministic
problem.

\section{Non-parametric quasi-Newton methods}
\label{sec:NPqN}

\subsection{A non-standard take on the quasi-Newton methods}
%
The idea underlying the Newton and quasi-Newton methods is that they
\emph{learn a local quadratic model} $q(x_k, \delta)$ of the cost
function $f(x)$ around the current iterate $x_k$
\begin{align}
  \label{eq:22}
  q(x_k,\delta) \triangleq f(x_k) + g(x_k)^{\Transp} \delta + \frac{1}{2} \delta^{\Transp} H(x_k) \delta,
\end{align}
where $\delta = x - x_k$, $g(x_k) = \nabla f(x) | _{x = x_k}$ denotes
the gradient and $H(x_k) = \nabla^2 f(x) | _{x = x_k}$ denotes the
Hessian. Note that~\eqref{eq:22} is a second order Taylor expansion of
$f(x)$, i.e.  $f(x) \approx q(x_k,\delta)$ in a close vicinity around
$x_k$. The quasi-Newton methods compute an estimate of the Hessian
based on zero and first order information (function values and their
gradients). More specifically these methods are designed to represent
the cost function according to the following model
\begin{align}
  \label{eq:16a}
  f_{\text{q}}(x_k+\delta) &= f(x_k) + g(x_k)^{\Transp}\delta + \frac{1}{2} \delta B_k \delta,
\end{align}
for some positive definite matrix $B_k$. Note that 
\begin{align}
  \label{eq:17a}
  \nabla_\delta f_{\text{q}}(x_k + \delta) &= g(x_k) + B_k \delta.
\end{align}
Quasi-Newton methods make a standing assumption that 
\begin{align}
  \label{eq:19a}
  \left . \nabla_\delta f_{\text{q}}(x_k + \delta) \right | _{\delta = x_{k-1} - x_{k}} = \nabla_x f(x_{k-1})
  = g(x_{k-1}). 
\end{align}
Equations~\eqref{eq:17a} and~\eqref{eq:19a} combined result in
$g(x_{k-1}) = g(x_k) + B_k (x_{k-1} - x_{k})$, so that if we define
\begin{align}
  \label{eq:21a}
  y_k \triangleq g(x_{k}) - g(x_{k-1}),\quad
  s_k \triangleq x_{k} - x_{k-1},
\end{align}
then we obtain the \emph{secant condition} (
\emph{quasi-Newton equation}),
\begin{align}
  \label{eq:22a}
  B_k s_k &= y_k.
\end{align}
This equation is not enough to define the elements of the Hessian
approximation $B_k$, we also know that by construction it has to be
symmetric. From a learning point of view this is very helpful since it
halves the number of unknown parameters to be estimated, but some
extra care will have to be taken to ensure this. That said, the
existing quasi-Newton algorithms can all be interpreted as employing
some particular form of regularisation on $B_k$, for example, one that
minimises changes from a previous Hessian approximation $B_{k-1}$. As
such, we can solve the following optimisation problem to find a
suitable $B_k$ as the solution to
\begin{equation}
  \label{eq:24}
  \begin{aligned}
    \min_{B} \quad &  \| B - B_{k-1} \|^2_W,\\
    \text{s.t.} \quad &  B=B^{\Transp},\\
    & Bs_k = y_k,
  \end{aligned}
\end{equation}
where $W$ is a positive definite weighting matrix. It was
Henning~\cite{HennigK:2013, Hennig:2015} who recently showed this
enlightening unifying interpretation of the quasi-Newton
algorithms. In Appendix~\ref{app:qNderivation} we also provide an
alternative derivation of the solution to~\eqref{eq:24} to complement
the developments in~\cite{Hennig:2015}. As pointed out
in~\cite{Hennig:2015} this opens up for some flexibility in finding
new algorithms which we will continue exploring below.

\subsection{Integral formulation of the quasi-Newton equation}
%
In the previous section we formulated the key quasi-Newton
equation~\eqref{eq:22a} using derivatives. We can represent the same
information using a \emph{line integral}, which comes about by noting
that if we define the line segment $r_k(\tau)$ between the current
iterate~$x_k$ and the previous iterate~$x_{k-1}$ as
\begin{align}
  \label{eq:46a}
  r_k(\tau) &\triangleq x_{k-1} + \tau(x_k-x_{k-1}), \qquad \tau \in [0,1],
\end{align}
then
\begin{align}
  \int_0^1 \frac{\partial }{\partial \tau} \nabla f(r_k(\tau)) \myd\tau
  \notag
  = \nabla f(r_k(1)) - \nabla f(r_k(0)) 
  = \nabla f(x_{k}) - \nabla f(x_{k-1})
 \triangleq y_k.
\end{align}
Now also note that by the chain rule
\begin{align}
  \label{eq:47a}
  \frac{\partial }{\partial \tau} \nabla f(r_k(\tau)) = \nabla^2
                                                        f(r_k(\tau))
                                                        \frac{\partial
                                                        r_k(\tau)}{\partial
                                                        \tau} 
  = \nabla^2 f(r_k(\tau)) (x_k - x_{k-1}).
\end{align}
Therefore,
\begin{align}
  \label{eq:48a}
   \int_0^1 \frac{\partial }{\partial \tau} \nabla f(r_k(\tau)) \myd\tau &= 
               \int_0^1 \nabla^2 f(r_k(\tau)) (x_k - x_{k-1}) \myd\tau.
\end{align}
That is 
\begin{align}
  \label{eq:49a}
  y_k =  \int_0^1 \nabla^2 f(r_k(\tau)) (x_k - x_{k-1}) \myd\tau.
\end{align}
This means that the difference between gradients (i.e. $y_k$) can be
considered a line integral observation of the Hessian
matrix. Therefore, in theory we could update an estimate of the
Hessian based on this and other such observations. However, since the
Hessian is unknown we do not have any functional form for
it. Hennig~\cite{Hennig:2013} introduced the idea of using a Gaussian
process to represent the true Hessian. This is the approach we will
take here as well. There are two key problems in building a Bayesian
non-parametric model of the Hessian using a GP. Firstly, we have to be
able to make use of the line integral observations~\eqref{eq:49a} when
learning the GP. This has been solved and used in other settings
before, see e.g. \cite{HennigK:2013,Wahlstrom:2015}. 
%
%
Secondly, how do we ensure that the resulting GP
represents a Hessian, i.e. that its realisations are at least
symmetric matrices? We will in the subsequent section develop a
solution based on the so-called \emph{half-vector
  operator}~\cite{MagnusN:1980} to ensure that the GP employed in
representing the Hessian is symmetric.  Importantly, the fact that the
gradient observations are potentially noisy does in fact not cause any
problems at all, since this fits within the standard Gaussian prosess
regression formulation.

\subsection{Modelling the Hessian as a GP}
%
The equivalent integral version of the quasi-Newton
equation~\eqref{eq:22a} was in the previous section shown to be
\begin{align}
  \label{eq:25}
  y_k = \int_0^1 B(r_k(\tau)) s_k \myd\tau,
\end{align}
where $B(\cdot)$ denotes the model of the Hessian and
$s_k = x_{k} - x_{k-1}$. Note that since $B(r_k(\tau)) s_k$ is a
column vector we can straightforwardly apply the vectorisation
operator inside the integral in~\eqref{eq:25} without changing the
result,
\begin{align}
  \label{eq:59a}
  y_k = \int_0^1 \vecV{ B(r_k(\tau)) s_k } \myd \tau 
  = \int_0^1 (s^{\Transp}_k \otimes I) \vecV{B(r_k(\tau))} \myd\tau 
  = (s^{\Transp}_k \otimes I) \int_0^1 \vecV{B(r_k(\tau))} \myd\tau,
\end{align}
where $\otimes$ denotes the Kronecker product. The whole point of this
exercise is that we have now isolated the vectorised Hessian estimate
$\vecV{B(r_k(\tau))}$ inside the integral. One option would now be to
place a GP prior on $\vecV{B(r_k(\tau))}$, but that would not enforce
the symmetry requirement we have on the Hessian estimate. We can solve
this problem using the half-vectorisation operator\footnote{For a
  symmetric $n\times n$ matrix $A$ the vector $\vecV{A}$ contains
  redundant information. More specifically, we do not need to keep the
  $n(n-1)/2$ entries above the main diagonal. The half-vectorisation
  $\vecVh{A}$ of a symmetric matrix $A$ is obtained by vectorising
  only the lower triangular part of $A$.}
$\vecVh{\cdot}$~\cite{MagnusN:1980}. Hence, we now assume that we have
a GP prior on the unique elements in the Hessian estimate
\begin{align}
  \label{eq:23}
  \widetilde{B}(\tau) = \vecVh{B(r_k(\tau))},
\end{align}
that is given by
\begin{align}
  \label{eq:61a}
  p\left(\widetilde{B}(\tau)\right) = \GP (\mu_k(\tau), \kappa_k(\tau,t)).
\end{align}
We can then retrieve the full Hessian estimate using the so-called
\emph{duplication matrix} $D$, which is a matrix such that  
\begin{align}
  \label{eq:28}
  \vecV{B(r_k(\tau))} = D \widetilde{B}(\tau).
\end{align}
More details and some useful results on the duplication matrix, the
associated \emph{elimination matrix} ($\vecVh{A} = L \vecV{A}$) and
their use are provided by~\cite{MagnusN:1980}. It is now
straightforward to also generalise the measurement~\eqref{eq:59a} by
adding some noise
\begin{align}
  \label{eq:60}
  y_k &= (s^{\Transp}_k \otimes I) \int_0^1 D   \widetilde{B}(\tau)
        \myd\tau + e_k, \quad e_k \sim \N(0, R).
\end{align}
It can now be shown that the joint GP for $\widetilde{B}(\tau)$ and
$y_k$ is given by
\begin{subequations}
  \begin{align}
    \label{eq:62}
    p(\widetilde{B}(\tau),y_k) = \GP \left (m_{\text{j}}, K_{\text{j}} \right ),
  \end{align}
  \begin{align}
    m_{\text{j}} = \bmat{\mu_k(\tau) \\ (s_k
    \otimes I) \int_0^1 \mu_k(\tau) \myd\tau}, \, K_{\text{j}} =     \bmat{\kappa_k(\tau,t) & \gamma_k(\tau,t)  \\
    \gamma^{\Transp}_k(\tau,t) & \pi_k(\tau,t) },
  \end{align}
\end{subequations}
where $\gamma_k(\tau,t)$ and $\pi_k(\tau,t)$ are given by
\begin{align}
  \label{eq:63}
  \gamma_k(\tau,t) &= \left (\int_0^1 \kappa_k(\tau,t) d\tau \right )
                     D^\Transp (s_k \otimes I),
\end{align}
and
\begin{align}
  \label{eq:64}
  \pi_k(\tau,t) = (s_k^{\Transp} \otimes I) D \left ( \int_0^1
  \int_0^1 \kappa_k(\tau,t)\, d\tau\, dt\right )D^\Transp (s_k \otimes I) + R.
\end{align}
Employing the standard results for conditioning of partitioned
Gaussians we obtain the posterior distribution
$p(\widetilde{B}(\tau)\mid y_k)$ from which we can then assemble back
the full Hessian estimate.
\begin{subequations}
  \label{eq:27a}
  \begin{align}
    p(\widetilde{B}(\tau)\mid y_k) = \GP(m, K),
  \end{align}
  where
  \begin{align}
    \label{eq:29}
    m &= \mu_k(\tau) +  \gamma_k(\tau,t)\pi^{-1}(\tau, t)(y_k -  (s_k \otimes I) \int_0^1 \mu_k(\tau) \myd\tau), \\
    K &= \kappa_k(\tau,t) - \gamma_k(\tau,t) \pi^{-1}(\tau, t) \gamma^{\Transp}_k(\tau,t).
  \end{align}
\end{subequations}
Finally, the Hessian estimate is according to~\eqref{eq:28} given by
\begin{align}
  \label{eq:26}
  p(\vecV{B(r_k(\tau))}\mid y_k) =  \GP(Dm, DKD^\Transp).
\end{align}

\subsection{Resulting optimisation algorithm}
The above ideas are collected here in the form of an algorithm
statement where the main theme is akin to quite standard
gradient-based search algorithms. In particular, we compute a search
direction based on gradient information and the Hessian approximation,
and perform a line search along this direction using the cost function
$f(x)$ to regulate a potential decrease in the cost. Importantly, care
must be taken when performing a line search in this setting since
$f(x)$ is stochastic. Here we employ the recent work
in~\cite{MahsereciH:2015} that delivers a line search algorithm
that handles noisy function and gradient evaluations and also satisfies
Wolfe-like conditions on the calculated step length.

It is important to be specific about the covariance
function employed below. Here we have opted to use a multi-variate
version of the squared exponential covariance given by
\begin{align}
  \label{eq:13}
  k_k(\tau,t) = \sigma^2 C_k e^{-\frac{1}{2}r_k^T(\tau) V r_k(t)}
\end{align}
where the matrix $C_k$ describes the covariance effect on each element
of $\widetilde{B}(\cdot)$, the matrix $V$ acts as an inverse length scale,
and $\sigma^2$ scales the entire covariance.
\begin{algorithm}[!hb]
\caption{\textsf{GP Hessian Approximation optimisation}}
\small
\begin{algorithmic}[1]
  \REQUIRE An initial estimate $x_1$ and a mean estimate of the Hessian matrix $\mu_1(\cdot)
  = \vecVh{B_1}$, and a covariance matrix $C_1$, and a positive
  integer $k_{\max} > 0$ that determines the maximum number of
  iterations. %
  \STATE Set $k = 1$ and perform the following.
  \WHILE{$k < k_{\max}$}
  \STATE Calculate a descent direction $p_k$ based on the current
  Hessian approximation $B_k$ and gradient $g(x_k)$ (care should be
  taken to ensure that this is a descent direction since $B_k$ is not
  guaranteed to be positive definite).
  \STATE Calculate a suitable step length $\alpha_k$ along the
  direction $p_k$ according to~\cite{MahsereciH:2015} and set $x_{k+1} = x_k + \alpha_k p$.
  \STATE Set $k \rightarrow k+1$.
  \STATE Update the Hessian approximation mean $\widetilde{B}_k = m$ and set the covariance
  matrix $C_k = K$ according to \eqref{eq:27a}.
  \ENDWHILE
\end{algorithmic}
\label{alg:GPhessianApprox}
\end{algorithm}

\input{global.tex}

%

\section{System identification experiments}
\label{sec:experiments}
As a testing ground for the probabilistic optimisation algorithms
developed and reviewed above we have chosen to study the problem of
identifying a nonlinear state-space model of the form
\begin{subequations}
  \begin{align}
    \label{eq:18}
    x_{t+1} &= f(x_t, \theta) + w_t,\\
    y_t &= g(x_t, \theta) + e_t.
  \end{align}
\end{subequations}
Note that we will in this section switch to the standard notation used
within system identification. Here $x_t\in \setX \subseteq \R^{\nX}$
and $y_t\in \setY \subseteq \R^{\nY}$ denotes the state and the
measurement, respectively. The dynamics and the measurements are
modeled by the nonlinear functions $f(\cdot)$ and $g(\cdot)$
parameterised by the unknown parameters
$\theta \in \setTh \subseteq \R^{\nTh}$. Finally, $w_t$ and $e_t$
denotes the process noise and measurement noise, respectively.

More specifically we will study the \emph{maximum likelihood}
formulation of the nonlinear system identification problem, which
amounts to finding a point estimate of the unknown parameter~$\theta$
in~\eqref{eq:18} by solving the following optimisation problem
\begin{align}
  \label{eq:20}
  \widehat{\theta}_{\text{ML}} = \arg \max_{\theta\in\Theta}{p_{\theta}(y_{1:\T}),}
\end{align}
where $y_{1:\T} = \{y_1, \dots, y_{\T}\}$. The likelihood function
$L(\theta) = p_{\theta}(y_{1:\T})$ is not available in closed form,
however using sequential Monte Carlo methods (a.k.a. particle
filters)~\cite{Gordon:1993,DoucetJ:2011} we can compute \emph{unbiased
  estimates} of the likelihood, by solving the following integral
\begin{align}
  \label{eq:31}
  L(\theta) = p_{\theta}(y_{1:\T}) = \int p_{\theta}(y_{1:\T}, x_{1:\T}) \myd x_{1:\T},
\end{align}
where the accuracy depends on the computational power we have
available. For recent overviews and links into the rapidly expanding
literature on the use of particle filters for nonlinear system
identification we refer to~\cite{SchonLDWNSD:2015,Kantas:2015}. The
idea of using a global GP model for the cost function in the nonlinear
system identification problem has previously been explored in
\cite{DahlinL:2014}, but only using noisy observations of the
Likelihood, not its gradients.

We will in Section~\ref{sec:NE:L} show the performance on a simple and
controlled example where we can compute true cost function and the
true optimal solution using alternative methods. This is to instill
confidence in that both methods do indeed perform as we expect them to
do on a simple example. In Section~\ref{sec:NE:NL} we will then study
a significantly harder nonlinear example.

\subsection{Simple Linear Example}
\label{sec:NE:L}
In order to gain some confidence in the probabilistic optimisation
methods, here we present the results of applying
Algorithms~\ref{alg:GPhessianApprox} and \ref{alg:GP} to a standard
linear state-space model identification problem. Specifically, we are
interested in estimating the parameters $\theta = \{a,c,q,r\}$ for the
following system
\begin{subequations}
  \label{eq:17}
  \begin{align}
    x_{t+1} &= a x_t + w_t, \qquad &w_t &\sim \mathcal{N}(0,q),\\
    y_t &= c x_t + e_t,  &e_t &\sim \mathcal{N}(0,r).
  \end{align} 
\end{subequations}
The true values for the system are $a^{\star}=0.9$, $c^{\star}=1.0$,
$q^{\star} = 0.1$ and $r^{\star} = 0.5$. The initial state is given by 
$x_1 \sim \mathcal{N}(0,1)$.

For a given set of measurements $y_{1:\T}$ it is possible to calculate
the likelihood $L(\theta)$ and its gradient $\nabla_\theta L(\theta)$
via standard Kalman filter equations and then employ standard
gradient-based search algorithms to obtain $\widehat{\theta}$ that
maximises the Likelihood.  In this regard, the problem does not suffer
from noisy Likelihood and gradient calculations, which serves the
purpose of profiling Algorithms~\ref{alg:GPhessianApprox} and
\ref{alg:GP} in the noise-free case.

To that end, we generated a Monte--Carlo simulation with $100$ runs,
where each run involves the generation of a new dataset $Y_N$
according to system \eqref{eq:17}. Furthermore, the initial parameter
vector $\theta_0$ was selected at random via moving each element
within a range of 50\% of the true value. A standard gradient-based
search algorithm and Algorithms~\ref{alg:GPhessianApprox} and
\ref{alg:GP} were all provided with the same initial conditions and
dataset for each Monte--Carlo run.

For this simulation study, the GP hyperparameters used in
Algorithm~\ref{alg:GPhessianApprox} were chosen as $B_1 = 100I$,
$C_1 = I$, $V = 10^{-3}I$ and $\sigma^2 = 1$. For
Algorithm~\ref{alg:GP} we employed a squared exponential covariance
function
$k(\theta,\theta^\prime) = \sigma^2
\exp(-0.5(\theta-\theta^\prime)^TV(\theta-\theta^\prime))$
with $\sigma = 200$ and $V$ chosen as a diagonal matrix with diagonal
entries $\{ 2, 2, 2, 20\}$.

\begin{figure}[!bth]
    \centering
    \begin{subfigure}[t]{.49\linewidth}
\centering        \includegraphics[width=0.9\columnwidth]{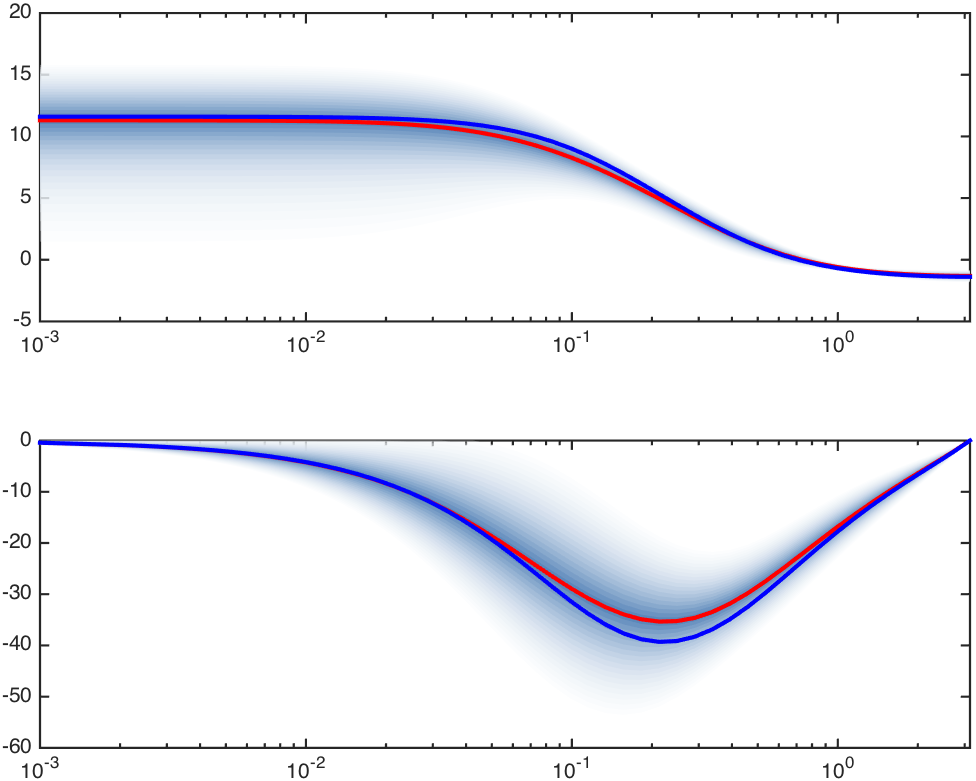}
       \caption{All three algorithms for noise free case on $L(\theta)$.}
        \label{fig:1}
      \end{subfigure}
      \begin{subfigure}[t]{.49\linewidth}
\centering        \includegraphics[width=0.9\columnwidth]{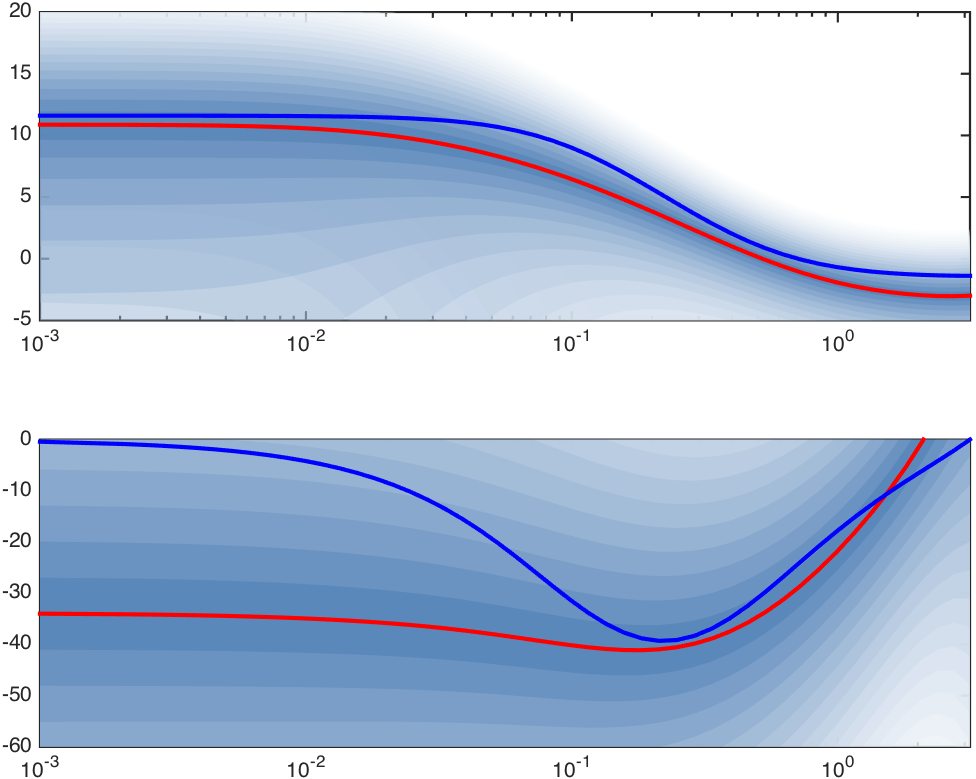}
       \caption{Classical BFGS algorithm for noisy observations of $L(\theta)$.}
        \label{fig:1}
      \end{subfigure}
      \begin{subfigure}[t]{.49\linewidth}
\centering        \includegraphics[width=0.9\columnwidth]{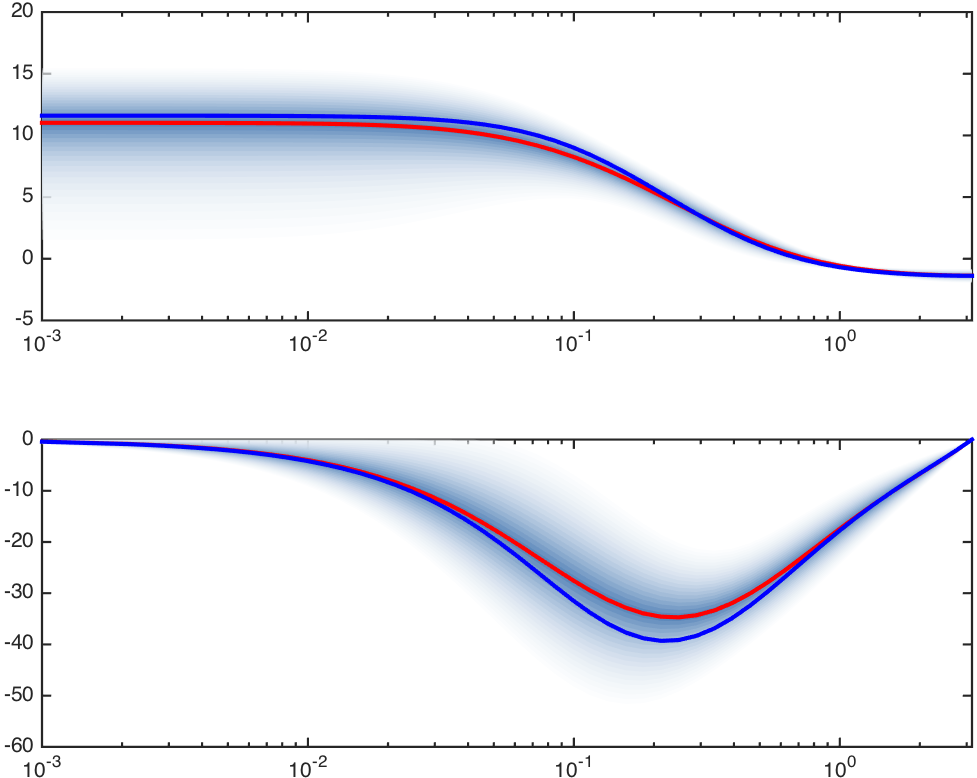}
        \caption{GP-based Hessian
          Approx. (Algorithm~\ref{alg:GPhessianApprox}) for noisy
          observations of $L(\theta)$.}
        \label{fig:2}
      \end{subfigure}
      \begin{subfigure}[t]{.49\linewidth}
\centering        \includegraphics[width=0.9\columnwidth]{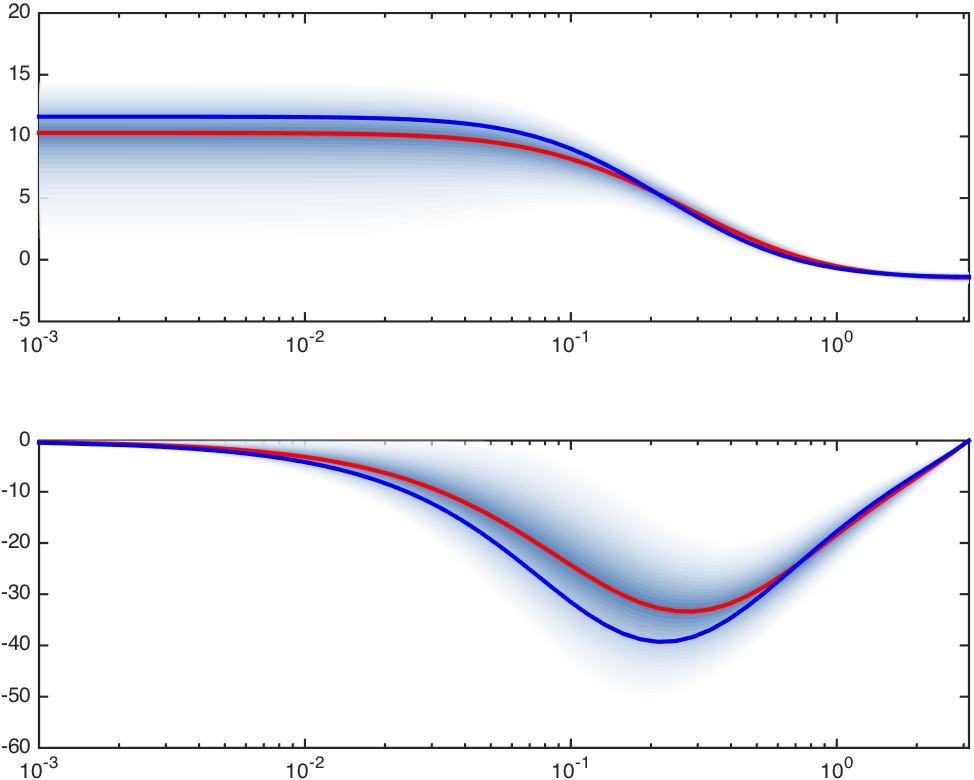}
        \caption{GP-based BFGS algorithm (Algorithm~\ref{alg:GP}) with
          noisy observations of $L(\theta)$.}
        \label{fig:2}
      \end{subfigure}
      \caption{Bode plots of estimated mean (light red) and true
        (blue) systems for cases of noise free (top--left), and
        noisy measurements (remaining plots). The blue shaded area
        represents the variability of the $100$ Monte-Carlo runs.}
      \label{fig:bode_no_noise}
\end{figure}

The top--left plot in Figure~\ref{fig:bode_no_noise} shows the
results for the Monte--Carlo runs. Perhaps not surprisingly, all
algorithms produced identical transfer function estimates for
the noise-free case, so we have shown only one plot.

Based on these positive results, we conducted a further Monte--Carlo
simulation, again comprising $100$ runs, where noise was deliberately 
added to both the Likelihood and gradient, Specifically, 
\begin{subequations}
  \label{eq:19}
  \begin{align}
    \widehat{L}(\theta) &= L(\theta) + v_c, \qquad &v_c &\sim
                          \mathcal(0, 10^4),\\
    \widehat{\nabla_\theta L}(\theta) &= \nabla_\theta L(\theta) +
                                        v_g,   &v_g &\sim
                                        \mathcal(0, 25 I ).
  \end{align}
\end{subequations}
Again, each run involved the generation of a new dataset and this time
the initial parameters were chosen as
$\theta_0 = \{a^{\star}/10, c^{\star}/10, q^{\star}/10,
r^{\star}/10\}$, in order to ensure that the results were not just a
function of randomly chosen initial parameters. Again, we ran both a
standard gradient-based search algorithm and 
Algorithms~\ref{alg:GPhessianApprox} and \ref{alg:GP} for each run.

The right-hand column of plots in Figure~\ref{fig:bode_no_noise} shows
the Bode responses for each estimated system. As possibly expected,
the standard gradient-search algorithm often fails to converge due to
the presence of noisy cost and gradient evaluations hence resulting in
a large variation of estimated transfer functions. In
many cases it is impossible to know if the search direction is
actually a 
descent direction, and at the same time a line-search algorithm often
fails to find a suitable scaling parameter since it is based on noisy
function evaluations.

Contrasting this, Algorithms~\ref{alg:GPhessianApprox} and
\ref{alg:GP} appear to generate estimates that have a similar
distribution to the noise-free case. It is difficult to discern which
of these two algorithms that performs best. One notable difference
between them is that Algorithm~\ref{alg:GP} was able to terminate
based on standard stopping criteria (small gradient norm for example),
which is made possible because the GP approximation is smooth. This is
not true of Algorithm~\ref{alg:GPhessianApprox}, which ran to the
maximum allowed (100) iterations for every run.

While we recognise that it is dangerous to draw definitive conclusions
from this limited study, it is nevertheless very encouraging results.

\subsection{More Challenging Nonlinear Example}
\label{sec:NE:NL}
Encouraged by the results obtained above for the noisy linear case,
here we consider a more challenging problem of identifying the
parameters $b$ and $q$ for the following nonlinear and time-varying
state-space model,
\begin{subequations}  
  \label{eq:nlp}
  \begin{align}
    x_{t+1} &= 0.5x_t + b\frac{x_t}{1 + x_t^2} + 8\cos (1.2t) + q w_t, \label{eq:54a}\\
    y_t &= 0.05x_t^2 + e_t,\label{eq:49}
  \end{align}
where
  \begin{align}
  \begin{bmatrix}
    w_t \\ 
    e_t
  \end{bmatrix} &\sim \mathcal{N} \left (\begin{bmatrix}
      0 \\ 
      0
    \end{bmatrix}, \begin{bmatrix}
      1 & 0 \\ 
      0 & 0.1
    \end{bmatrix}
  \right)
  \label{eq:55a}
\end{align}
\end{subequations}
and the true parameters are $b^\star = 25$ and $q^\star = 0.1^{1/2}$. 
This example has previously been investigated by the current
authors~\cite{SchonWN:2011} and is profiled again here due to it being
acknowledged as a challenging 
problem~\cite{DoucetGA:2000,GodsillDW:2004}.

Algorithms~\ref{alg:GPhessianApprox} and \ref{alg:GP} were employed to
estimate $b$ and $q$ based on $100$ Monte--Carlo runs using $N=100$
data points for each run as generated by~\eqref{eq:nlp} with the true
parameter values. The initial parameter values were chosen randomly in
each simulation where the value was chosen uniformly within a $50\%$
range of the true value. The algorithms were allowed to iterate for no
more than 100 iterations.

In this case, Algorithm~\ref{alg:GPhessianApprox} was employed with
exactly the same hyperparameter choices as for the linear
example. Algorithm~\ref{alg:GP} was employed using the covariance
function used above, but with $\sigma=10^3$ and $V$ chosen as a
diagonal matrix with entries $\{0.01, 1\}$.

In the case of Algorithm~\ref{alg:GP} we employed the availability of
the Hessian approximation from the GP in order to employ a Newton type
algorithm, rather than a quasi-Newton algorithm as before. While this
is not strictly necessary, it highlights the flexibility of that
approach.

The likelihood and its gradient cannot be calculated exactly in this
case and we therefore employed sequential Monte Carlo methods and
Fisher's identity \cite{CappeMR:2005,NinnessWS:2010} to provide noisy
estimates of both. The number of particles used to calculate these
terms was $500$ in all cases. Note that each simulation required no
more than 8 seconds of computation time on a MacBook Pro 2.8GHz Intel
i7.

\begin{figure}[!bth]
    \centering
      \begin{subfigure}[t]{.49\linewidth}
        \includegraphics[width=\columnwidth]{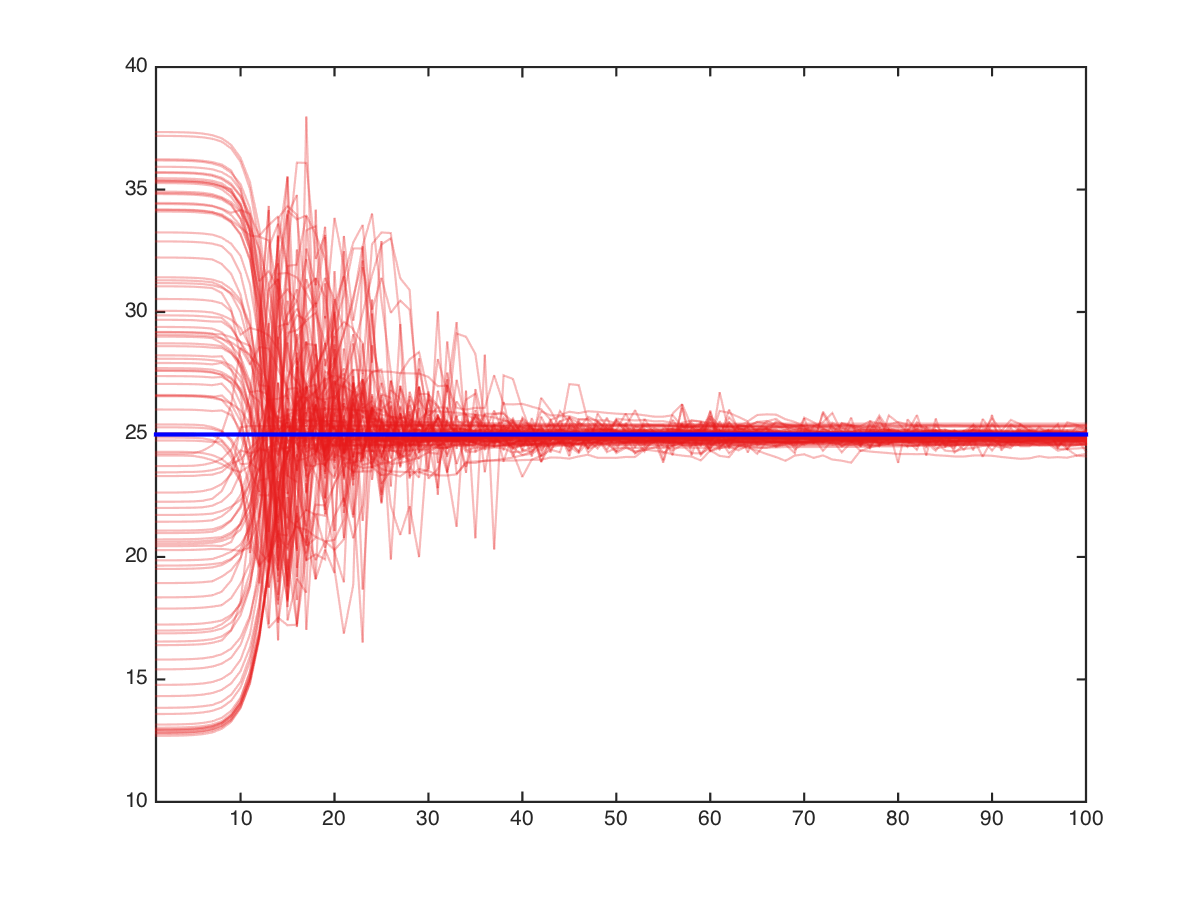}
        \caption{Iterations for~$b$ using 
          the Hessian approx. in
          Algorithm~\ref{alg:GPhessianApprox}.}
        \label{fig:nl2}
      \end{subfigure}
      \begin{subfigure}[t]{.49\linewidth}
        \includegraphics[width=\columnwidth]{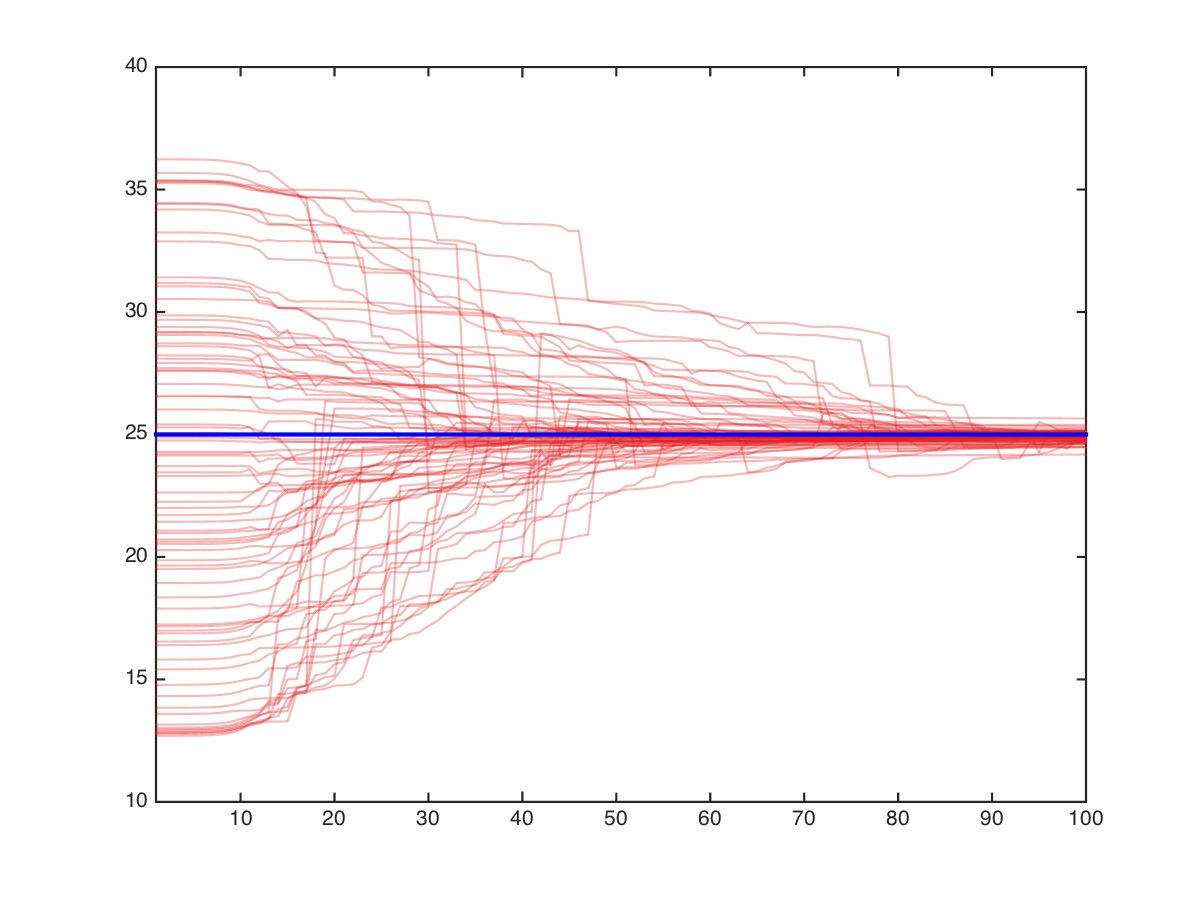}
        \caption{Iterations for~$b$ using the global GP from
          Algorithm~\ref{alg:GP}.}
        \label{fig:nl1}
      \end{subfigure}
      \begin{subfigure}[t]{.49\linewidth}
        \includegraphics[width=\columnwidth]{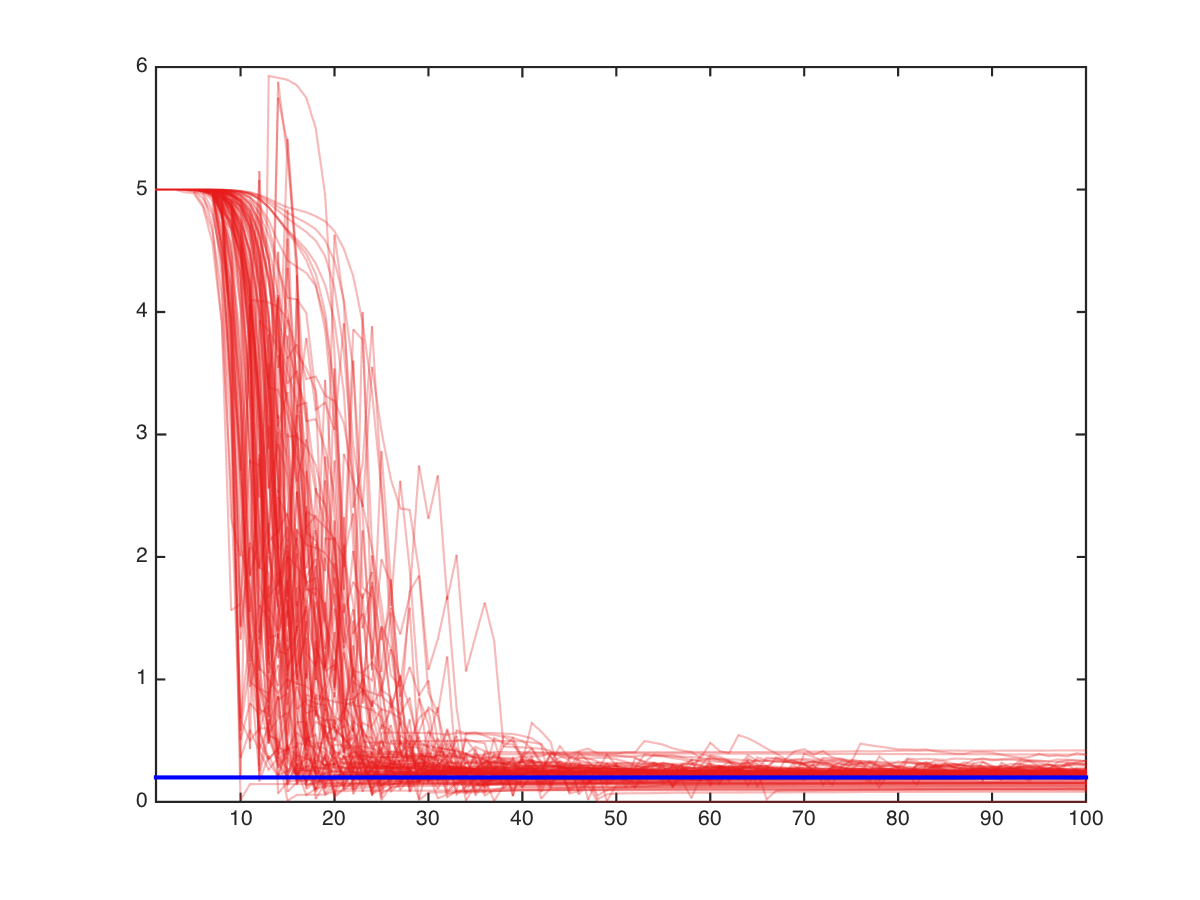}
        \caption{Iterations for~$q$ using 
          the Hessian approx. in
          Algorithm~\ref{alg:GPhessianApprox}.}
        \label{fig:nl4}
      \end{subfigure}
      \begin{subfigure}[t]{.49\linewidth}
        \includegraphics[width=\columnwidth]{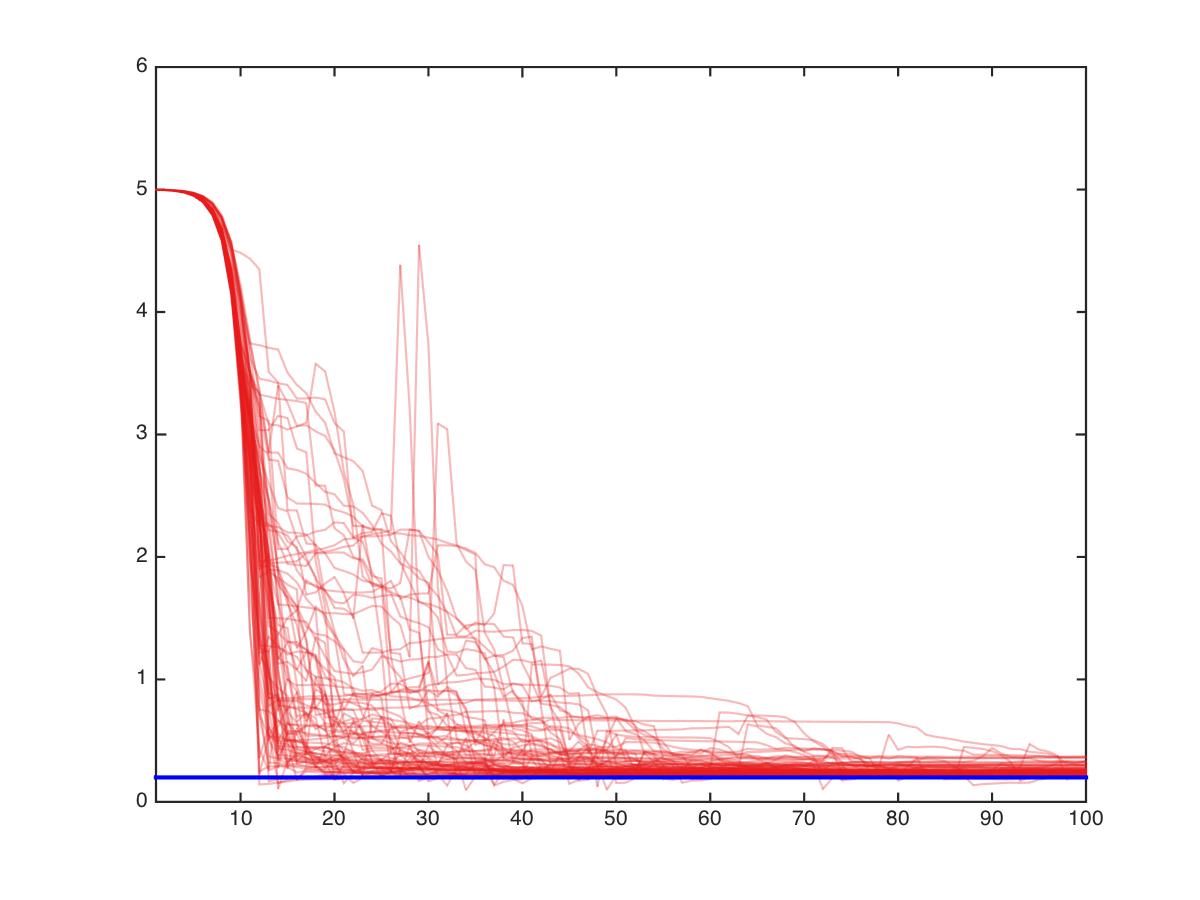}
        \caption{Iterations for~$q$ using the global GP from
          Algorithm~\ref{alg:GP}.}
        \label{fig:nl3}
      \end{subfigure}
      \caption{Iterations of parameter values using the GP Hessian
        approximation in Algorithm~\ref{alg:GPhessianApprox} (left
        column) and the global GP of Algorithm~\ref{alg:GP} (right
        column) with estimates in red and true value shown as solid
        blue.}
      \label{fig:nonlin}
\end{figure}

The results of this Monte--Carlo simulation can be observed in
Figure~\ref{fig:nonlin}. For Algorithm~\ref{alg:GP} we have removed 19
of the 100 simulation results due to convergence to a local minima,
which resulted in a final parameter value that was greater than 5\%
in error relative true parameter value. For
Algorithm~\ref{alg:GPhessianApprox} we removed only 1 simulation
result due to the same criteria. 

Again, it is dangerous to draw too many conclusions from these
results. At the same time, the performance of
Algorithm~\ref{alg:GPhessianApprox} appears to be slightly better than 
Algorithm~\ref{alg:GP}, which may be related to the choice of
hyperparameters for the latter method.

\begin{figure}[!bth]
    \centering
    \begin{subfigure}[t]{.49\linewidth}
        \includegraphics[width=\columnwidth]{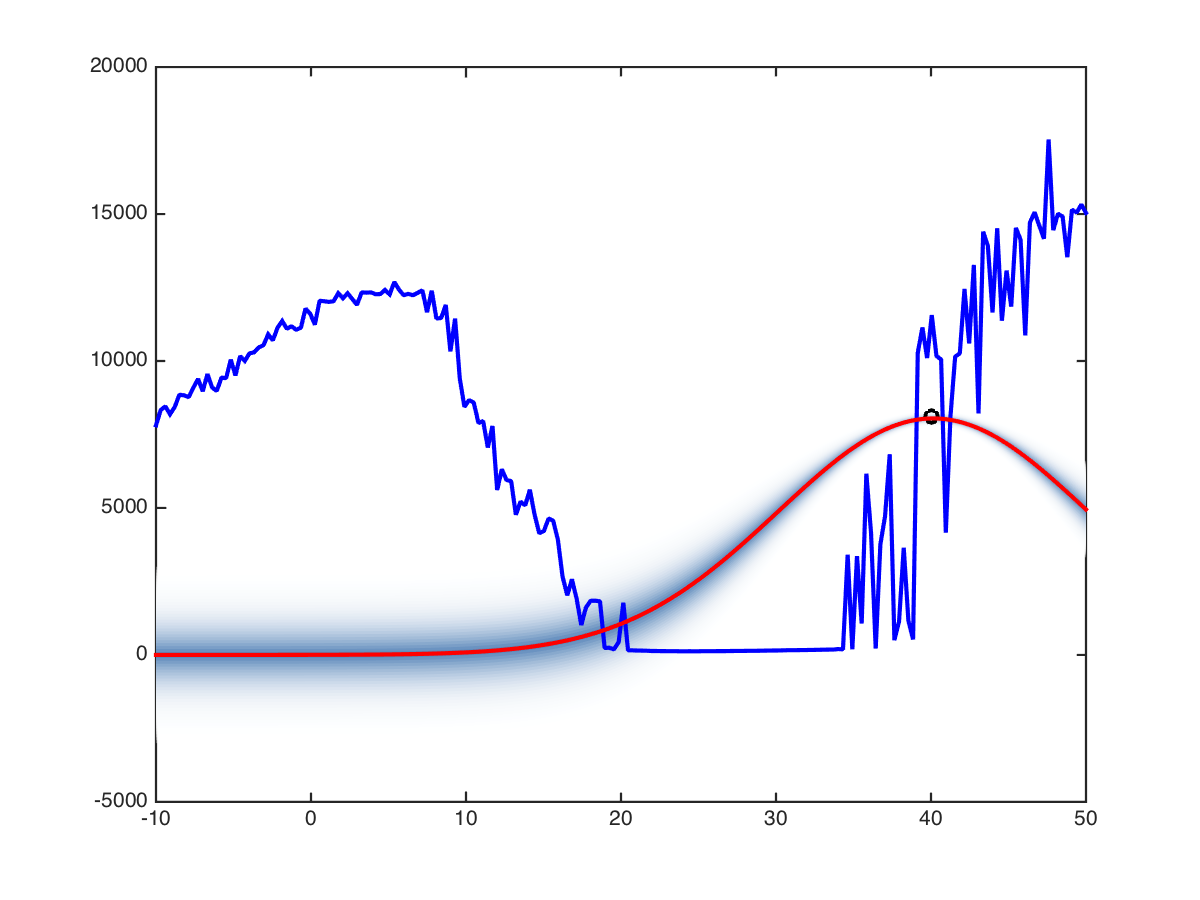}
       \caption{GP after 1 sample.}
        \label{fig:nl1}
      \end{subfigure}
      \begin{subfigure}[t]{.49\linewidth}
        \includegraphics[width=\columnwidth]{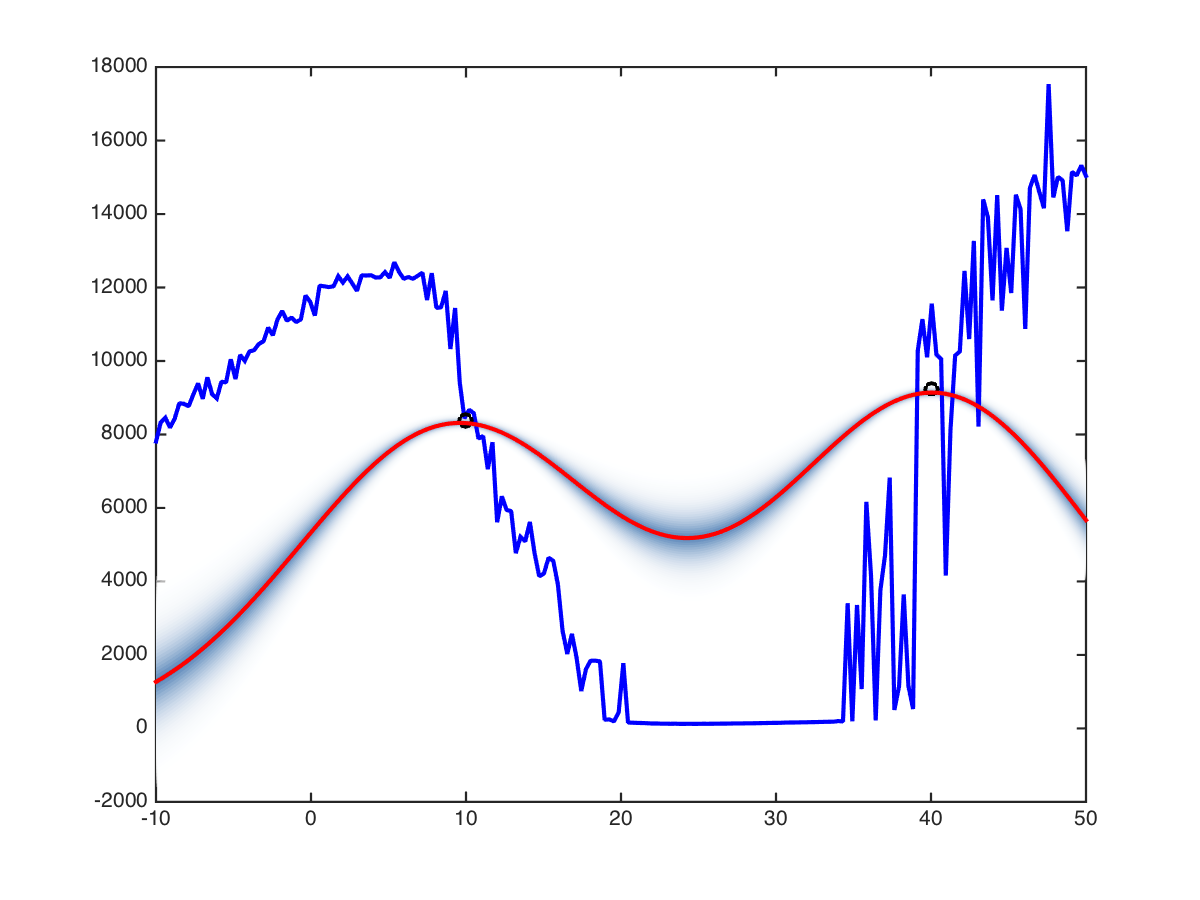}
        \caption{GP after 2 samples.}
        \label{fig:nl2}
      \end{subfigure}
      \begin{subfigure}[t]{.49\linewidth}
        \includegraphics[width=\columnwidth]{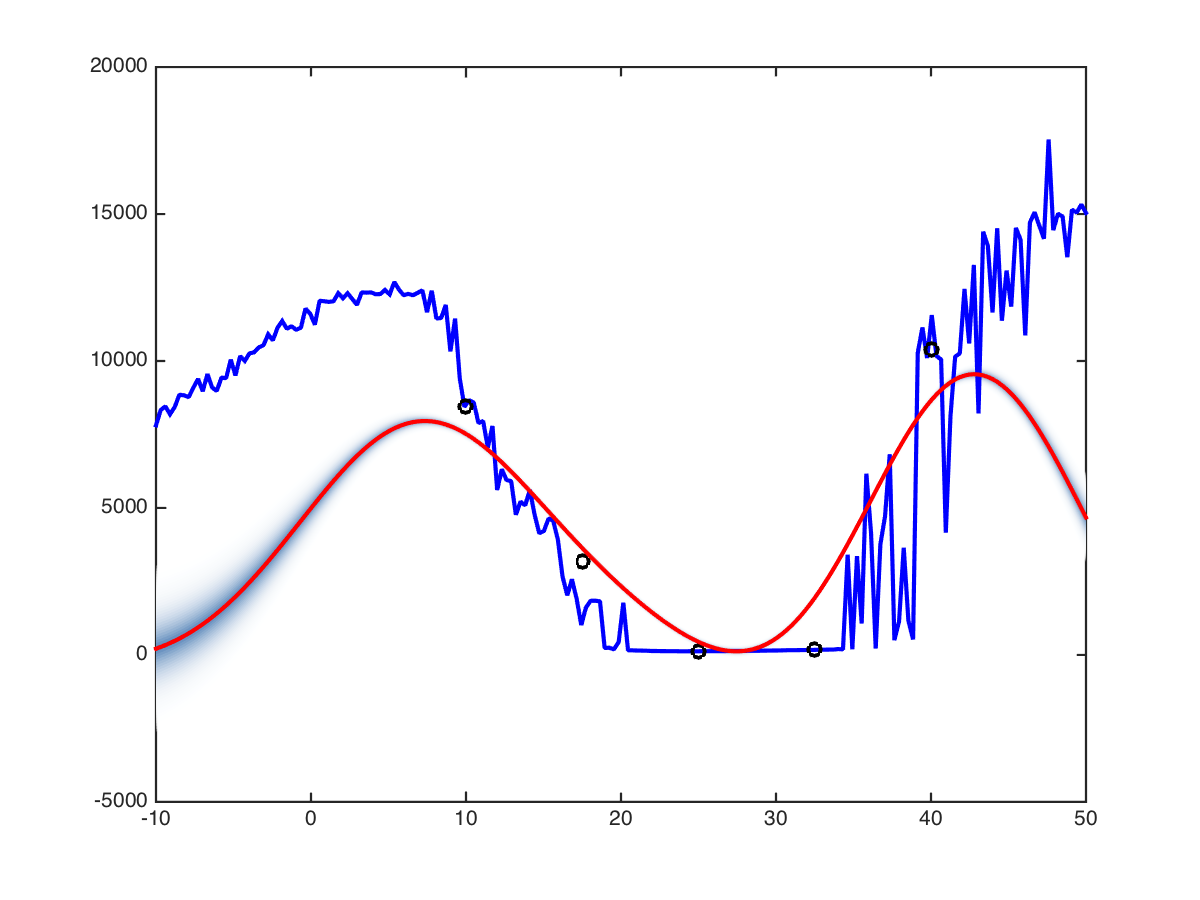}
       \caption{GP after 5 samples.}
        \label{fig:nl3}
      \end{subfigure}
      \begin{subfigure}[t]{.49\linewidth}
        \includegraphics[width=\columnwidth]{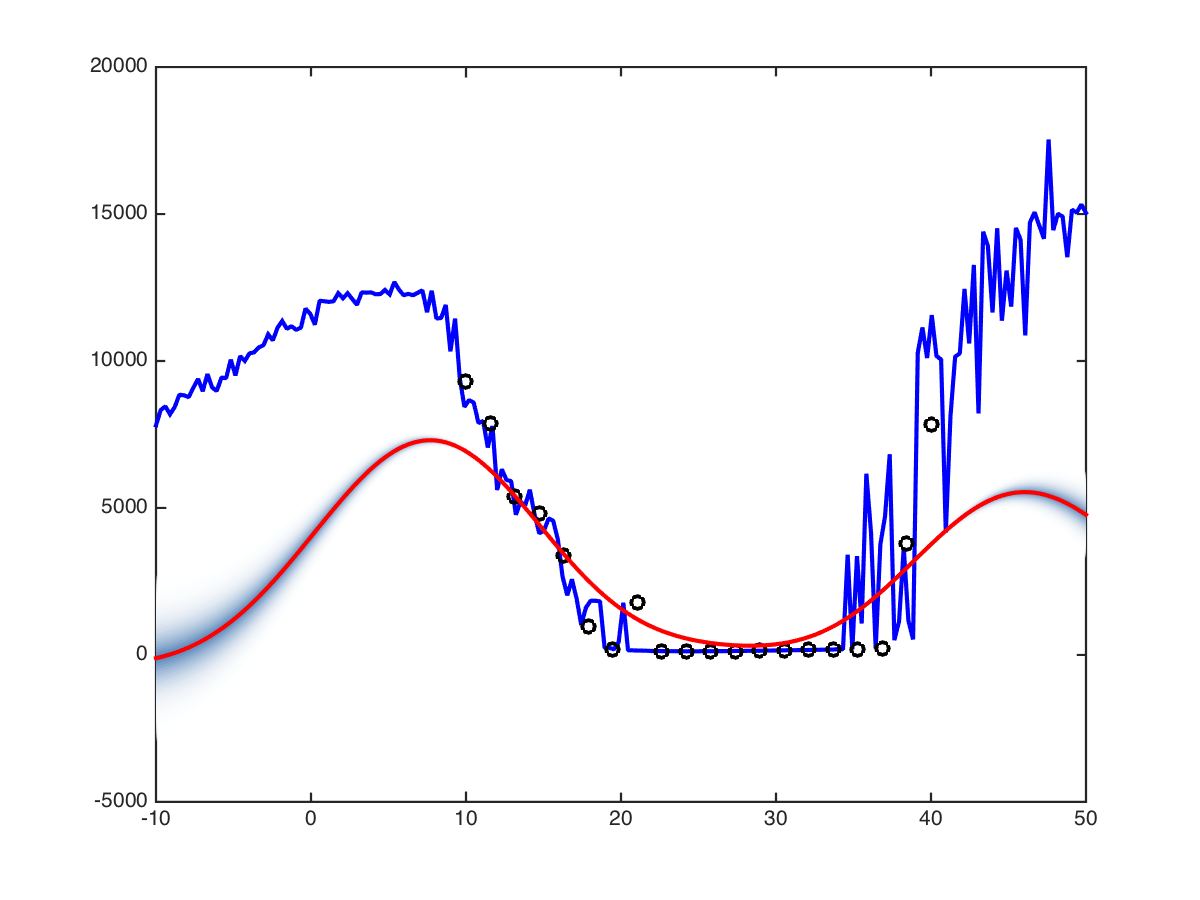}
        \caption{GP after 20 samples.}
        \label{fig:nl4}
      \end{subfigure}
      \caption{The surrogate cost function modeled as a global GP
        according to Algorithm~\ref{alg:GP} (red solid line with grey
        shading to indicate uncertainty) and the true cost function
        given by the likelihood (solid blue). Sample points are shown
        as black circles.}
      \label{fig:nonlin_gp_approx}
\end{figure}

It is interesting to visually observe the effect of using the global
GP model offered by Algorithm~\ref{alg:GP} in terms of smoothing the
cost function. The current example is known to exhibit erratic
likelihood behaviour at extreme points in the parameter
space~\cite{NinnessWS:2010}. This is perhaps best visualised by
restricting to just one parameter, in this case the $b$
parameter. Figure~\ref{fig:nonlin_gp_approx} shows a sequence of plots
where progressively more samples are used to model the underlying
likelihood. It can be verified in these plots that the likelihood
changes rapidly at the extremities of the $b$ range, and yet the GP
approximation remains smooth and captures the global behaviour.

\section{Discussion}\label{sec:conclusions}
Minimising a nonlinear cost function $f(x)$ is a challenging problem
in general, and as verified again here, is made even more difficult if
the cost function and its derivatives cannot be evaluated without
unknown errors. These latter types of stochastic problems have been
considered for some time now, and yet very recent results
in~\cite{ByrdHNS:2016} show that this is still an active area of
research. The main thrust of current activities is to capture the
curvature information available from noisy gradient measurements. In
the current paper, we have developed a new approach and reviewed an
existing approach for capturing this curvature information that both
rely on Bayesian non-parametric estimates of the unknown functions. The
first treats the Hessian matrix as an unknown function and employs
integral observations of the gradient vector in order to form the
curvature estimate. The second approach employs a GP for the entire
cost function and is based on noisy observations of the cost, its
gradient and possibly even Hessian matrix.

Both approaches appear to have merit and we believe that these
approaches deserve further attention. Specifically, to the best of our
knowledge, the choice of covariance functions and the corresponding
selection of hyperparameters has not been explored in a rigorous
manner. For example, Student-t processes~\cite{ShahWG:2014} might be a
natural way to reduce the effect of spurious large errors in the
function or gradient. The question of adaptively tuning the
hyperparameters as these algorithms progress also deserves more
attention. Other areas to explore include suitable stopping criteria
for problems involving stochastic cost functions, and the
problem of reducing computational load by
employing GP approximations.

\section{Acknowledgements}
This research was financially supported by the projects
\emph{Probabilistic modeling of dynamical systems} (Contract number:
621-2013-5524), and \emph{NewLEADS - New Directions in Learning
  Dynamical Systems} (Contract number: 621-2016-06079), both funded by
the Swedish Research Council and the project \emph{ASSEMBLE} (Contract
number: RIT15-0012) funded by the Swedish Foundation for Strategic
Research (SSF).

\appendix
 \section{Derivation of the standard quasi-Newton equations}
\label{app:qNderivation}
Recall the optimisation problem outlined in~\eqref{eq:24},
\begin{align}
  \label{eq:24b}
  B_k &= \min_{B} \| B - B_{k-1} \|^2_W \quad \textnormal{s.t.} \quad B=B^{\Transp} \quad \textnormal{and} \quad Bs_k = y_k.
\end{align}
Note that 
\begin{align}
  \label{eq:25}
  \| B - B_{k-1} \|^2_W \triangleq \textnormal{trace}\
  W(B-B_{k-1})^{\Transp} W(B - B_{k-1}) 
   = (\vv{B} - \vv{B}_{k-1}) ^{\Transp} (W \otimes W) (\vv{B} - \vv{B}_{k-1})  
\end{align}
and where $\vv{B}$ is the usual vectorization operator that stacks the columns of $B$ on top of one another to form a column vector, and $\otimes$ is the Kronecker product.

Another way to write the optimisation problem is in terms of the
column vector $\vv{B}$ so that $\vv{B}_k$ is the solution to the
following constrained optimization problem
\begin{equation}
  \begin{aligned}
    \max_{\vv{B}} \quad &  (\vv{B} - \vv{B}_{k-1}) ^{\Transp} (W \otimes W) (\vv{B} - \vv{B}_{k-1}), \\
    \text{s.t.} \quad &\Gamma \vv{B} = 0,\\
    & (s_k^{\Transp} \otimes I) \vv{B} = y_k,
  \end{aligned}
\end{equation}
where $\Gamma \triangleq I - P$, and $P$ is the vec-permutation matrix that has the property $P\vv{B} = \vv{B^{\Transp}}$. 
Therefore, $\Gamma\vv{B} = 0$ is equivalent to $B = B^{\Transp}$.
If we define some new variables
\begin{subequations}
  \begin{align}
    b &\triangleq \vv{B},\\
    b_k &\triangleq \vv{B_k},\\
    V &\triangleq W \otimes W,\\
    S_k &\triangleq s_k^{\Transp} \otimes I,
  \end{align}
\end{subequations}
then we can rewrite the minimisation problem as 
\begin{equation}
  \label{eq:28}
  \begin{aligned}
    b_k = \min_{b}  \quad &  (b - b_{k-1}) ^{\Transp} V (b - b_{k-1}), \\
    \text{s.t.} \quad &\Gamma b = 0,\\
    &S_k b = y_k.
  \end{aligned}
\end{equation}
A suitable Lagrangian for this problem is
\begin{align}
  \label{eq:29}
  L(b,\lambda,\eta) = \frac{1}{2}b^{\Transp} Vb - b^{\Transp} Vb_{k-1} + \frac{1}{2}b_{k-1}^{\Transp} V b_{k-1} - \lambda^{\Transp} (\Gamma b) + \eta^{\Transp} (y_k-S_kb).
\end{align}
Note that the first order necessary conditions of optimality require that
\begin{align}
  \label{eq:30}
  \nabla_b L &= V b - Vb_{k-1} - \Gamma^{\Transp} \lambda - S_k^{\Transp} \eta = 0.
\end{align}
This implies that 
\begin{align}
  \label{eq:31}
  \Gamma^{\Transp} \lambda = V(b-b_{k-1}) - S_k^{\Transp} \eta.
\end{align}
Recall that $\Gamma = \Gamma^{\Transp}$ and that if we define 
\begin{align}
  \label{eq:33}
  \lambda &\triangleq \vec{\Lambda},
\end{align}
then
\begin{align}
  \label{eq:32}
  \Gamma^{\Transp} \lambda = \vv{\Lambda - \Lambda^{\Transp}}.
\end{align}
So from \eqref{eq:31} and the inverse vec operator
\begin{align}
  \label{eq:34}
  \Lambda - \Lambda^{\Transp} &= W(B - B_{k-1})W - \eta s_k^{\Transp}.
\end{align}
We can add $(\Lambda - \Lambda^{\Transp})^{\Transp}$ to the left hand side to reveal that 
\begin{align}
  \label{eq:35}
  \Lambda - \Lambda^{\Transp} + (\Lambda - \Lambda^{\Transp})^{\Transp} = 0,
\end{align}
so that
\begin{align}
  \label{eq:36}
  0 &=  W(B - B_{k-1})W - \eta s_k^{\Transp} + W(B - B_{k-1})W - s_k \eta^{\Transp}.
\end{align}
In the above we have exploited the fact that $W$, $B$ and $B_{k-1}$ are symmetric. 
This implies that 
\begin{align}
  \label{eq:37}
  B &= B_{k-1} + \frac{1}{2}W^{-1}(\eta s_k^{\Transp} + s_k \eta^{\Transp})W^{-1}.
\end{align}
From the constraints we have that 
\begin{align}
  \label{eq:38}
  Bs_k &= y_k = B_{k-1}s_k + \frac{1}{2}W^{-1}(\eta s_k^{\Transp} + s_k \eta^{\Transp})W^{-1}s_k.
\end{align}
So that
\begin{align}
  \label{eq:39}
  W^{-1}(\eta s_k^{\Transp} + s_k \eta^{\Transp})W^{-1}s_k &= 2(y_k - B_{k-1}s_k).
\end{align}
This implies that 
\begin{align}
  \label{eq:7}
    \eta &= \frac{2W(y_k - B_{k-1}s_k) - s_k \eta^{\Transp} W^{-1}s_k}{s_k^{\Transp} W^{-1}s_k}
\end{align}
Post multiplying $\eta^{\Transp}$ by $W^{-1}s_k$ results in
\begin{align}
  \notag
  \eta^{\Transp} W^{-1}s_k &= \frac{2(y_k - B_{k-1}s_k)^{\Transp} W W^{-1}s_k - s_k^{\Transp} W^{-1} \eta s_k^{\Transp} W^{-1}s_k}{s_k^{\Transp} W^{-1}s_k}\\
  \notag
                   &= \frac{2(y_k - B_{k-1}s_k)^{\Transp} W W^{-1}s_k}{s_k^{\Transp} W^{-1}s_k} - s_k^{\Transp} W^{-1} \eta\\
                   &= \frac{2(y_k - B_{k-1}s_k)^{\Transp} W W^{-1}s_k}{s_k^{\Transp} W^{-1}s_k} - \eta^{\Transp} W^{-1} s_k
\end{align}
Therefore
\begin{align}
  \label{eq:9}
  2\eta^{\Transp} W^{-1}s_k &= \frac{2(y_k - B_{k-1}s_k)^{\Transp} W W^{-1}s_k}{s_k^{\Transp} W^{-1}s_k}
\end{align}
Cancelling the common $2$ factor and substituting this into \eqref{eq:7} provides
\begin{align}
  \label{eq:10}
  \eta = \frac{2W(y_k - B_{k-1}s_k) - s_k \frac{(y_k - B_{k-1}s_k)^{\Transp} s_k}{s_k^{\Transp} W^{-1}s_k}}{s_k^{\Transp} W^{-1}s_k}
       = \frac{2W(y_k - B_{k-1}s_k)}{s_k^{\Transp} W^{-1}s_k} - \frac{s_k (y_k - B_{k-1}s_k)^{\Transp} s_k}{(s_k^{\Transp} W^{-1}s_k)^2}
\end{align}
Recall from \eqref{eq:37} that 
\begin{align}
  \label{eq:15}
    B &= B_{k-1} + \frac{1}{2}W^{-1}(\eta s_k^{\Transp} + s_k \eta^{\Transp})W^{-1}
\end{align}
Substituting \eqref{eq:10} into \eqref{eq:15} results in
\begin{align}
  \label{eq:40}
  B = B_{k-1} &+ \frac{(y_k - B_{k-1}s_k)s_k^{\Transp} W^{-1}}{s_k^{\Transp} W^{-1}s_k} - \frac{1}{2} \frac{W^{-1} s_k (y_k - B_{k-1}s_k)^{\Transp} s_k s_k^{\Transp} W^{-1}}{(s_k^{\Transp} W^{-1}s_k)^2}\nonumber\\
&+ \frac{W^{-1}s_k(y_k - B_{k-1}s_k)^{\Transp}}{s_k^{\Transp} W^{-1}s_k} - \frac{1}{2} \frac{W^{-1} s_k s_k^{\Transp} (y_k - B_{k-1}s_k) s_k^{\Transp} W^{-1}}{(s_k^{\Transp} W^{-1}s_k)^2}
\end{align}
Noting that $s_k^{\Transp} (y_k - B_{k-1}s_k) = (y_k - B_{k-1}s_k)^{\Transp} s_k$ and collecting like terms gives
\begin{align}
  \label{eq:41}
  B = B_{k-1} &+ \frac{W^{-1}s_k(y_k - B_{k-1}s_k)^{\Transp}  + (y_k - B_{k-1}s_k)s_k^{\Transp} W^{-1}}{s_k^{\Transp} W^{-1}s_k} - \frac{W^{-1} s_k (y_k - B_{k-1}s_k)^{\Transp} s_k s_k^{\Transp} W^{-1}}{(s_k^{\Transp} W^{-1}s_k)^2}
\end{align}

\bibliographystyle{ieeetr}

\end{document}

%% file: global.tex
\section{Global GP representation}
\label{sec:GlobalGP}
The approach adopted here is to optimise a surrogate function, rather
than the cost function itself. It is important that the surrogate
function maintain the global ``shape'' of the underlying cost function
and at the same time remain amenable to classical optimisation methods
for smooth functions. 

As a potential surrogate function, here we employ the non-parametric
class of GPs to model the cost function, its gradient and Hessian,
similar to the development in~\cite{OsborneGR:2009}. In particular, we
model the cost function $f(x)$ via
\begin{align}
  \label{eq:1}
  f(x) \sim \GP (\mu(x), k(x,x^\prime),
\end{align}
where $\mu(x)$ is some suitable mean function (for example, a
strictly convex function centred on prior knowledge of the parameter
values). Let us now introduce the notation
\begin{align}
  g(x) = \nabla_x f(x), \qquad
  h(x) = \vecVh{\nabla^2_x f(x)},
\end{align}
for the gradient and the Hessian, respectively. Here we have again
explicitly exploit the fact that the Hessian is symmetric and employ
the half-vectorisation operator to this end. Recall that the
derivative of a Gaussian process is another Gaussian
process~\cite{RasmussenW:2006}. Hence, if the covariance function
$k(x,x^\prime)$ is twice differentiable, the stacked object
$(f(x), g^{\Transp}(x), h^{\Transp}(x))^{\Transp}$ is guaranteed to be
a Gaussian process with induced mean and covariance functions
according to
\begin{align}
  \label{eq:2}
  \bmat{f(x) \\ 
  g(x) \\ 
  h(x)} \sim \GP \left ( \bmat{1 \\ \nabla_x \\
  \widetilde{\nabla}^2_x} \mu(x),  \bmat{1 \\ \nabla_x \\
  \widetilde{\nabla}^2_x} k(x,x^\prime) \bmat{1 \\ \nabla_{x^\prime} \\ \widetilde{\nabla}^2_{x^\prime}}^{\Transp}\right )
\end{align}
In the above, $\nabla_x$ is used to present the vector of partial
derivatives with respect to $x$ and
$\widetilde{\nabla}^2_x = \vecVh{\nabla^2_x}$ is used to represent the
vector of operators that is formed by applying the half-vectorization
operator to the following matrix of second order derivative operators
\begin{align}
  \label{eq:3}
  \nabla^2_x \triangleq \bmat{\frac{\partial^2}{\partial
  x_1 \partial x_1} & \frac{\partial^2}{\partial
  x_2 \partial x_1} & \cdots & \frac{\partial^2}{\partial
  x_n \partial x_1} \\ \frac{\partial^2}{\partial
  x_1 \partial x_2} & \frac{\partial^2}{\partial
  x_2 \partial x_2} & \cdots & \frac{\partial^2}{\partial
  x_n \partial x_2} \\ 
  \vdots & \vdots & \ddots & \vdots\\
  \frac{\partial^2}{\partial x_1 \partial x_n} & \frac{\partial^2}{\partial
  x_2 \partial x_n} & \cdots & \frac{\partial^2}{\partial
  x_n \partial x_n}}
\end{align}
Therefore, $\nabla^2_x \in \R^{n \times n}$ and
$\widetilde{\nabla}^2_x \in \R^{n(n+1)/2 \times 1}$.

For a given value of the parameters $x$, we obtain the following noisy
measurements of the cost function~$f(x)$, its gradient~$g(x)$ and its
Hessian~$h(x)$,
\begin{subequations}
  \label{eq:4}
  \begin{align}
    \label{eq:4a}
    \widehat{f}(x) &= f(x) + v_c, \quad &v_c &\sim \mathcal{N}(0,\sigma_c^2), \\
    \label{eq:4b}
    \widehat{g}(x) &= g(x) + v_g, &v_g &\sim \mathcal{N}(0,\Sigma_g),\\
    \widehat{h}(x) &= h(x) + v_h, &v_h &\sim \mathcal{N}(0,\Sigma_h),
  \end{align}
\end{subequations}
where $\sigma_c$, $\Sigma_g$ and $\Sigma_h$ carries information about
the nature of the noise. In a situation where we do not have access to
one or more of the observations in~\eqref{eq:4} we simply just remove
the corresponding line(s) and our development still holds. 

Based on the above it is now possible to construct a joint GP
consisting of both the true objects $\ell(x)$ and the (possibly noisy)
observations of them $\widehat{\ell}(x)$
\begin{align}
  \label{eq:7}
  \ell(x) &= \bmat{f(x) \\ 
  g(x) \\
  h(x)}, \qquad \widehat{\ell}(x) = \bmat{\widehat{f}(x) \\ \widehat{g}(x)  \\ \widehat{h}(x)},
\end{align}
according to
\begin{align}
  \label{eq:6}
  \bmat{\ell(x) \\ \widehat{\ell}(x)} \sim \GP \left (
  \bmat{\mu_\ell(x) \\ \mu_{\widehat{\ell}}(x)} ,
  \bmat{k_{\ell,\ell}(x,x^\prime) &
                                              k_{\ell,\widehat{\ell}}(x,x^\prime)
  \\ k_{\widehat{\ell},\ell}(x,x^\prime) & k_{\widehat{\ell},\widehat{\ell}}(x,x^\prime)}
  \right )
\end{align}
where
\begin{align}
  \label{eq:8}
   \mu_{\ell}(x) &= \bmat{\mu(x) \\ \nabla_x \mu(x) \\
  \widetilde{\nabla}^2_x \mu(x)}, \qquad 
  \mu_{\widehat{\ell}}(x) = \bmat{\widehat\mu(x) \\ 
  \nabla_x \widehat\mu(x)\\ 
  \widetilde{\nabla}^2_x \widehat\mu(x)}
\end{align}
and finally,
\begin{align}
  \label{eq:9}
  k_{\ell,\ell}(x,x^\prime) =  k_{\ell,\widehat{\ell}}(x,x^\prime)
  = \bmat{k(x,x^\prime)
                                        & k(x,x^\prime)
                                          \nabla_{x^\prime}^{\Transp} & k(x,x^\prime)
                                          (\widetilde{\nabla}^2_{x^\prime})^{\Transp}
                                                                     \\
  \nabla_x k(x,x^\prime)
                                        & \nabla_x k(x,x^\prime)
                                          \nabla_{x^\prime}^{\Transp} &
                                                                     \nabla_x  k(x,x^\prime)
                                          (\widetilde{\nabla}^2_{x^\prime})^{\Transp}
                                                                     \\
\widetilde{\nabla}^2_x k(x,x^\prime)
                                        & \widetilde{\nabla}^2_x k(x,x^\prime)
                                          \nabla_{x^\prime}^{\Transp} & \widetilde{\nabla}^2_x k(x,x^\prime)
                                          (\widetilde{\nabla}^2_{x^\prime})^{\Transp}
}
\end{align}
and $k_{\widehat{\ell},\ell}(x,x^\prime) =
k_{\ell,\widehat{\ell}}(x,x^\prime)^{\Transp}$, 
\begin{align}
  \label{eq:11}
  k_{\widehat{\ell},\widehat{\ell}}(x,x^\prime) =
  k_{\ell,\ell}(x,x^\prime) + \bmat{\sigma^2_c & 0 & 0\\
  0 & \Sigma_g & 0\\
  0 & 0 & \Sigma_h}.
\end{align}
The utility of this model is that if we have a collection of
observations $\{\widehat{\ell}(x_1), \ldots , \widehat{\ell}(x_N) \}$
then we can infer the cost function (and its associated gradient and
Hessian) at \emph{any} $x$ value based on the standard conditional
formulas
\begin{align}
  \label{eq:12}
  \ell(x) & = \mu_\ell(x) + K_{\ell,\widehat{\ell}}
                 K^{-1}_{\widehat{\ell},\widehat{\ell}}\bmat{\widehat{\ell}(x_1) - \mu_\ell(x_1) \\ \vdots
  \\ \widehat{\ell}(x_N) - \mu_\ell(x_N)},
\end{align}
where
\begin{align}
  \label{eq:14}
  K_{\widehat{\ell},\widehat{\ell}}  &=
  \bmat{k_{\widehat{\ell},\widehat{\ell}}(x_1,x_1) &
                                                               k_{\widehat{\ell},\widehat{\ell}}(x_1,x_2)
  & \cdots & k_{\widehat{\ell},\widehat{\ell}}(x_1,x_N) \\
k_{\widehat{\ell},\widehat{\ell}}(x_2,x_1) &
                                                       k_{\widehat{\ell},\widehat{\ell}}(x_2,x_2)
  & \cdots & k_{\widehat{\ell},\widehat{\ell}}(x_2,x_N) \\ 
\vdots & \vdots & \ddots & \vdots \\
k_{\widehat{\ell},\widehat{\ell}}(x_N,x_1) &
                                                       k_{\widehat{\ell},\widehat{\ell}}(x_N,x_2)
  & \cdots & k_{\widehat{\ell},\widehat{\ell}}(x_N,x_N)}\\
  \label{eq:15}
    K_{{\ell},\widehat{\ell}}  &=
  \bmat{k_{{\ell},\widehat{\ell}}(x,x_1) &
                                                               k_{{\ell},\widehat{\ell}}(x,x_2)
  & \cdots & k_{{\ell},\widehat{\ell}}(x,x_N)}.
\end{align}
Therefore, the function $\ell(x)$ in \eqref{eq:12} can be considered a
surrogate for the true cost function~$f(x)$, its gradient~$g(x)$ and
its Hessian~$h(x)$. Importantly, the smoothness properties of
$\ell(x)$ are now controlled by the choice of which covariance
function $k(\cdot,\cdot)$ to use, which acts as a filtering
mechanism. This is critical, since it allows us to make use of
standard optimisation routines for smooth cost functions in
optimising~$\ell(x)$. For example, classical Newton's method
algorithms (or quasi-Newton methods) can be employed. Importantly,
observations of the cost function and its gradient that are collected
during the search procedure can be rolled into the model so that
$\ell(x)$ more accurately models $f(x)$. These ideas are formalised in
Algorithm~\ref{alg:GP}.

\begin{algorithm}[!t]
\caption{\textsf{GP gradient-based optimisation}}
\small
\begin{algorithmic}[1]
  \REQUIRE A termination threshold value $\epsilon >0$.
  \STATE Set $k=1$, and initialise $x_1$.
  \STATE Based on~$x_1$, estimate the cost function noise
  covariance $v_c$ and the gradient covariance $\Sigma_g$ empirically.
  \WHILE{$\| g(x_k) \| > \epsilon$}
  \STATE Obtain the cost and gradient measurements
  $\widehat{\ell}(x_k)$.
  \STATE Use a gradient-based search algorithm to minimise
  $f(x)$ from $\ell(x)$ in~\eqref{eq:12} and set $x_{k+1}$ to be the
  minimising argument. Note: $\ell(x)$ must be
  updated for each new iterate generated within the search algorithm.
  \STATE Set $k\rightarrow k+1$.
  \ENDWHILE
\end{algorithmic}
\label{alg:GP}
\end{algorithm}

\begin{figure}[!hbt]
    \centering
    \begin{subfigure}[t]{.49\linewidth}
        \includegraphics[width=\columnwidth]{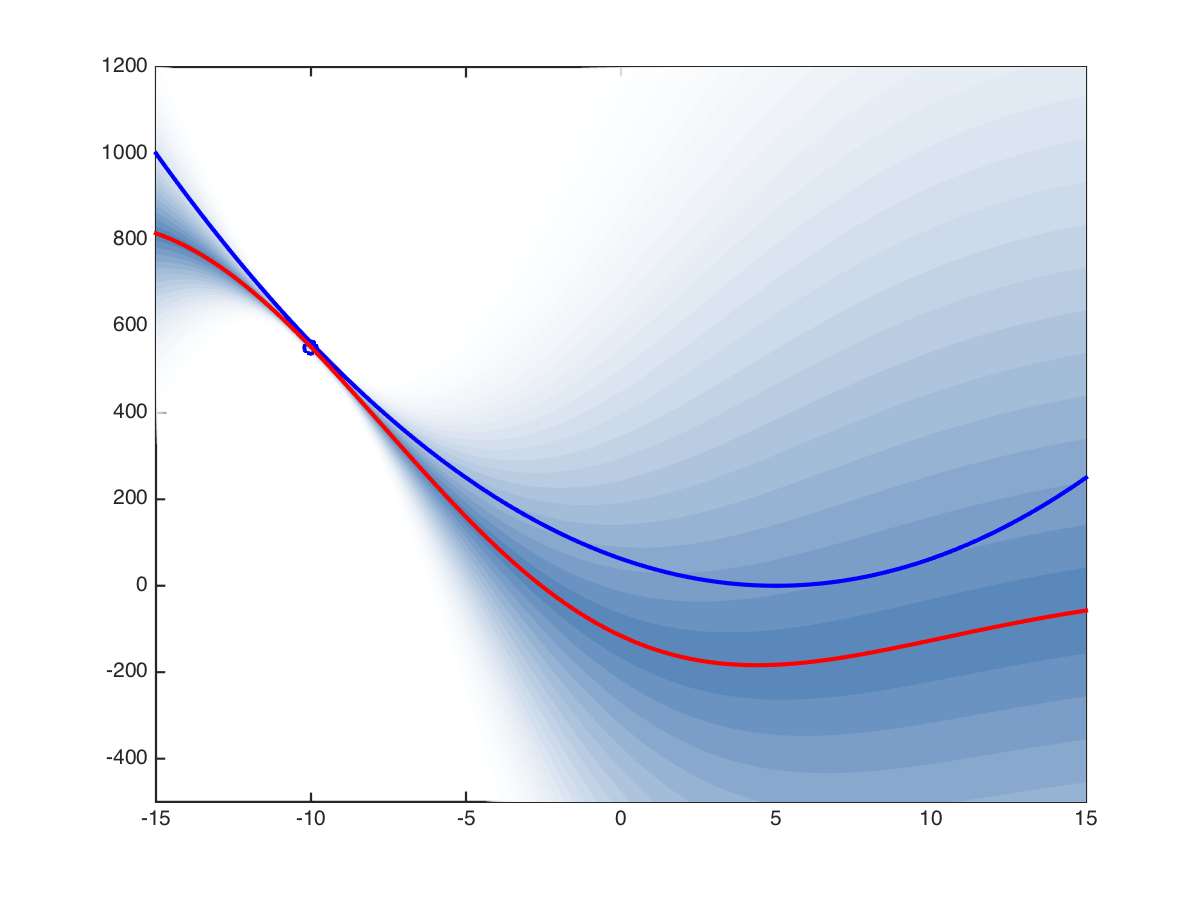}
       \caption{Using 1 observation.}
        \label{fig:1}
      \end{subfigure}
      \begin{subfigure}[t]{.49\linewidth}
        \includegraphics[width=\columnwidth]{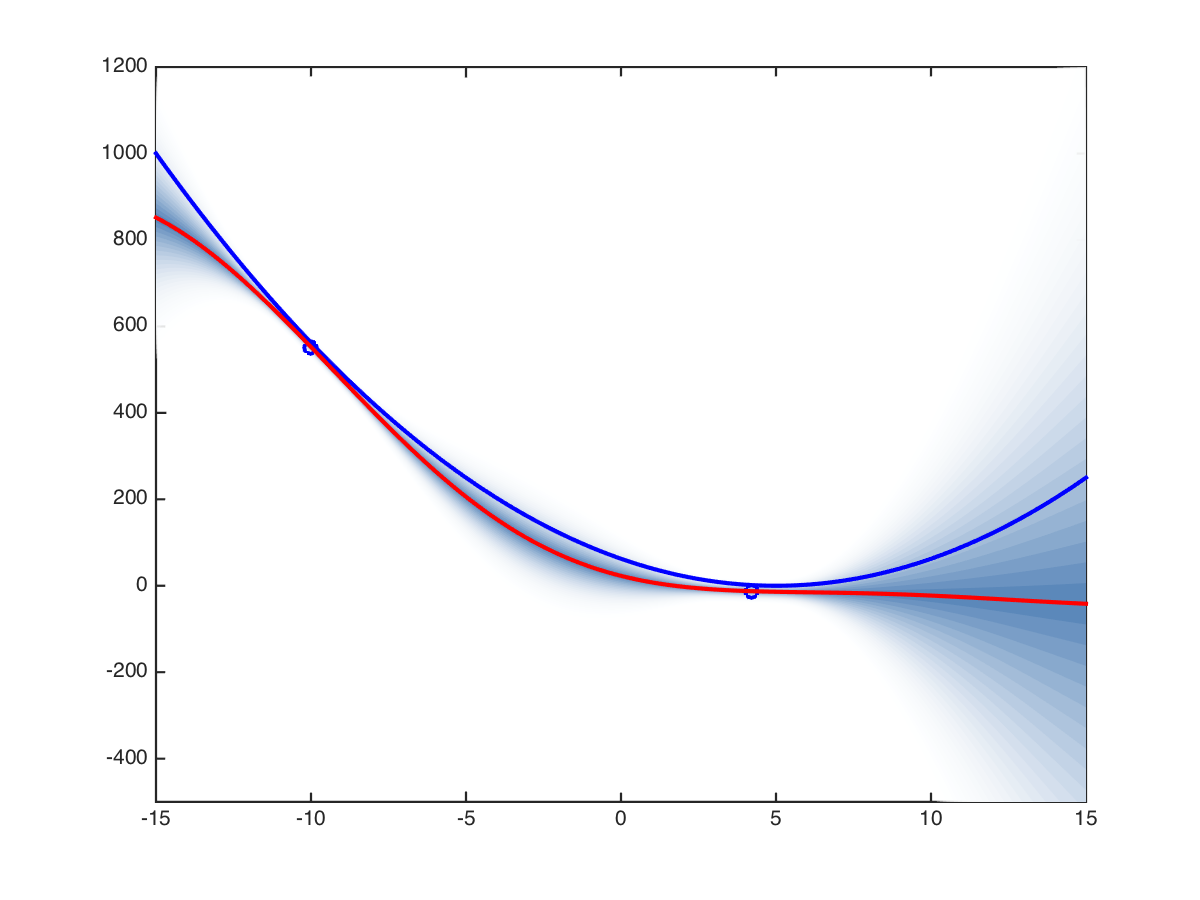}
        \caption{Using 2 observations.}
        \label{fig:2}
      \end{subfigure}
      \begin{subfigure}[t]{.49\linewidth}
        \includegraphics[width=\columnwidth]{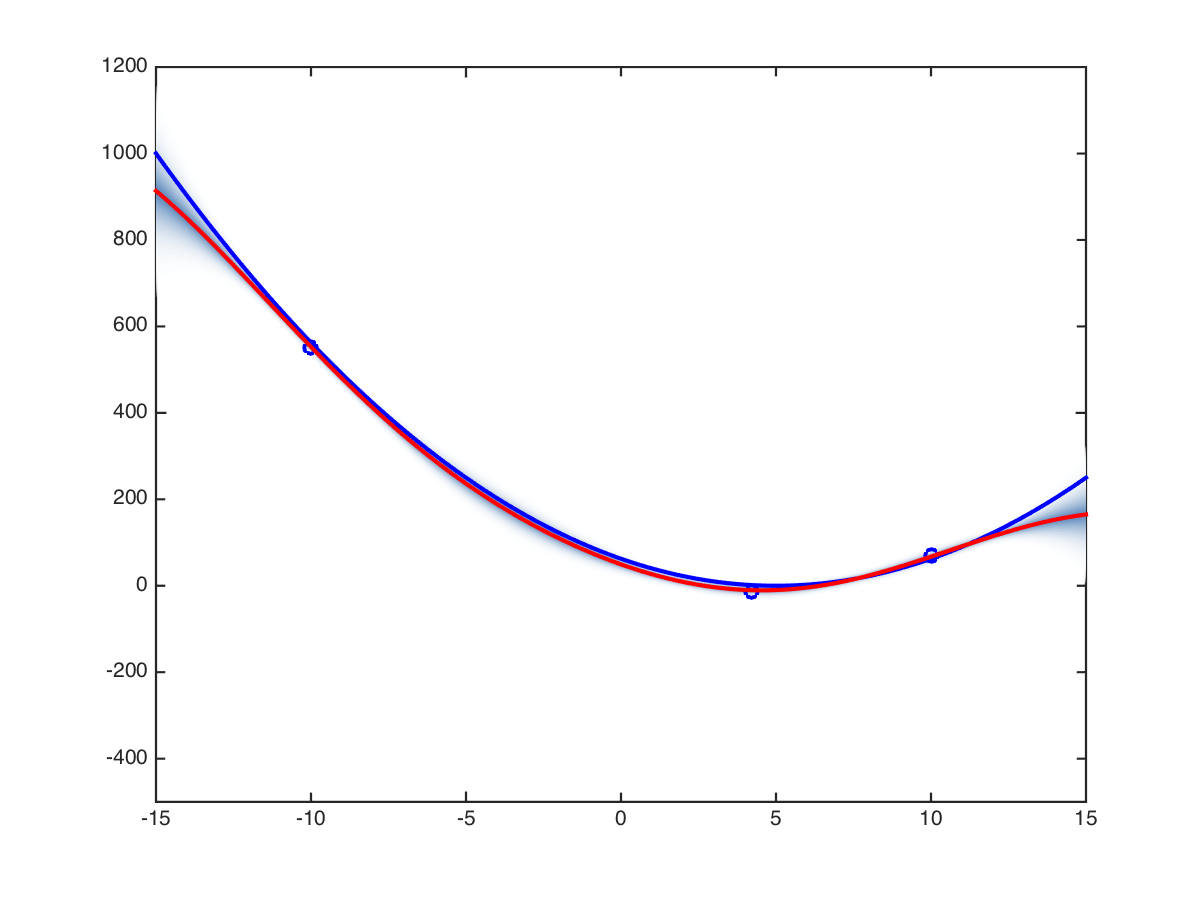}
      \caption{Using 3 observations.}
        \label{fig:3}
      \end{subfigure}
      \begin{subfigure}[t]{.49\linewidth}
        \includegraphics[width=\columnwidth]{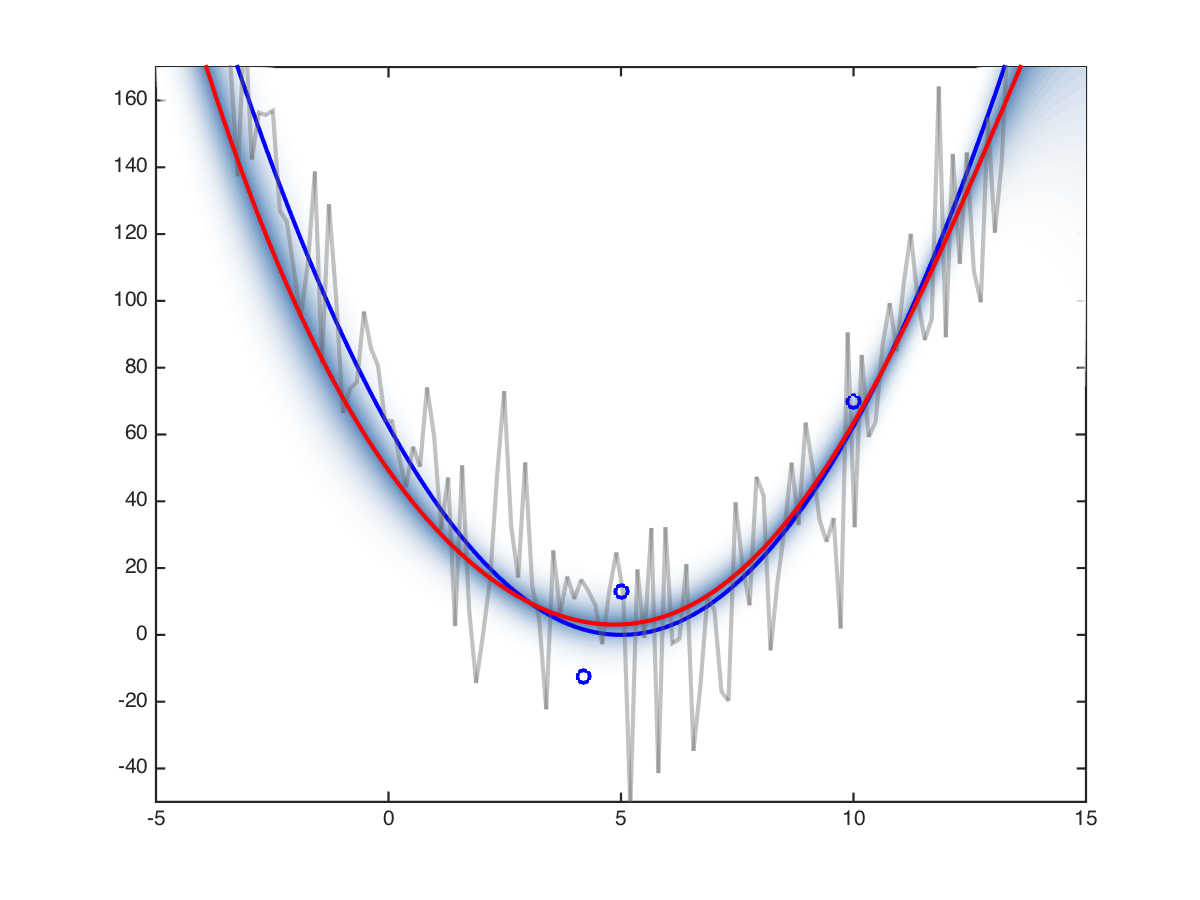}
        \caption{Using 4 observations.}
        \label{fig:4}
      \end{subfigure}
      \caption{GP approximation to a quadratic cost function. True
        function (blue), GP mean (red), uncertainty (shaded blue),
        noisy function value (blue circle). Figure (d) also shows the
        noisy cost function (light gray). Note that Figure (d) is
        zoomed.}
      \label{fig:123_all}
\end{figure}

By way of a pedagogical example, consider a very simple problem where
the cost function is given by the following quadratic function
\begin{align}
  \label{eq:5}
  f(x) = \frac{5}{2} (x - 5)^2, \qquad x \in \R.
\end{align}
Assume that we only have access to noisy measurements of the cost
function and its gradient according to~\eqref{eq:4a}--\eqref{eq:4b}
with $\sigma_c = 20$ and $\Sigma_g = 1$. Assuming that the mean
function~$\mu_\ell(x)$ is zero everywhere, then
Figure~\ref{fig:123_all} shows
the true cost function and the GP estimate as a standard
gradient-based search algorithm proceeds based on the surrogate
$\ell(x)$ starting at $x_1 = -10$. Here, we have used a
squared-exponential kernel
$k(x,x^\prime) = \sigma^2 e ^{-0.5l(x-x^\prime)^{2}}$, with
$\sigma = 10^3$ and $l = 0.01$. This sequence of plots
shows that the GP approach manages to capture the global shape of the
cost function, while at the same time remaining largely un-affected by
the noise. More interesting and challenging examples are provided in
the subsequent section.
